\documentclass[10pt,twocolumn,letterpaper]{article}

\usepackage[pagenumbers]{cvpr} %

\usepackage[dvipsnames]{xcolor}

\usepackage{tikz}
\usepackage{pgfplots}
\usepackage{calc}
\usepackage{mathtools}
\usepackage{amsfonts}
\usepackage{amssymb}
\usepackage{amsthm}
\usepackage{tabularx}
\usepackage{listings}
\usepackage{comment}
\usepackage{multirow}
\usepackage{xurl} %
\usepackage{comment}

\usepackage{multirow}
\usepackage{makecell}
\usepackage{colortbl}

\usetikzlibrary{spy}
\usetikzlibrary{shapes.arrows}
\usetikzlibrary{arrows.meta, arrows}
\usetikzlibrary{patterns, shapes.arrows}
\usetikzlibrary{positioning}
\usetikzlibrary{matrix}
\usetikzlibrary{pgfplots.groupplots}
\usepgfplotslibrary{groupplots, dateplot}
\usepgfplotslibrary{fillbetween}
\tikzset{every picture/.style={/utils/exec={\sffamily\fontsize{7}{8}}}}
\pgfplotsset{compat=1.10}

\newcommand \colorindicator[2]{%
	#1~{(\textcolor{#2}{$\blacksquare\!\!\!\!\!\blacksquare$})}%
}

\newcommand \colorindicatorn[2]{%
	#1~{\textcolor{#2}{$\blacksquare\!\!\!\!\!\blacksquare$}}%
}

\pgfplotsset{
  every axis plot/.append style={line width=0.7pt},
}

\definecolor{tud0d}{cmyk/RGB/HTML}{0,0,0,.8/83,83,83/535353}
\definecolor{tud0c}{cmyk/RGB/HTML}{0,0,0,.6/137,137,137/898989}
\definecolor{tud0b}{cmyk/RGB/HTML}{0,0,0,.4/181,181,181/B5B5B5}
\definecolor{tud0a}{cmyk/RGB/HTML}{0,0,0,.2/220,220,220/DCDCDC}
\definecolor{tud1a}{cmyk/RGB/HTML}{.7,.4,0,0/93,133,195/5D85C3}
\definecolor{tud2a}{cmyk/RGB/HTML}{0.8,.2,0,0/0,156,218/009CDA}
\definecolor{tud3a}{cmyk/RGB/HTML}{0.7,0,.5,0/80,182,149/50B695}
\definecolor{tud4a}{cmyk/RGB/HTML}{.4,0,.8,0/175,204,80/AFCC50}
\definecolor{tud5a}{cmyk/RGB/HTML}{.2,0,.8,0/221,223,72/DDDF48}
\definecolor{tud6a}{cmyk/RGB/HTML}{0,.1,.7,0/255,224,92/FFE05C}
\definecolor{tud7a}{cmyk/RGB/HTML}{0,.3,.8,0/248,186,60/F8BA3C}
\definecolor{tud8a}{cmyk/RGB/HTML}{0,.6,.8,0 /238,122,52/EE7A34}
\definecolor{tud9a}{cmyk/RGB/HTML}{0,.8,.7,0/233,80,62/E9503E}
\definecolor{tud10a}{cmyk/RGB/HTML}{.2,.9,0,0/201,48,142/C9308E}
\definecolor{tud11a}{cmyk/RGB/HTML}{.6,.8,0,0/128,69,151/804597}
\definecolor{tud1b}{cmyk/RGB/HTML}{1,.6,0,0/0,90,169/005AA9}
\definecolor{tud2b}{cmyk/RGB/HTML}{1,.3,0,0/0,131,204/0083CC}
\definecolor{tud3b}{cmyk/RGB/HTML}{1,0,.6,0/0,157,129/009D81}
\definecolor{tud4b}{cmyk/RGB/HTML}{.5,0,1,0/153,192,0/99C000}
\definecolor{tud5b}{cmyk/RGB/HTML}{.3,0,1,0/201,212,0/C9D400}
\definecolor{tud6b}{cmyk/RGB/HTML}{0,.2,1,0/253,202,0/FDCA00}
\definecolor{tud7b}{cmyk/RGB/HTML}{0,.4,1,0/245,163,0/F5A300}
\definecolor{tud8b}{cmyk/RGB/HTML}{0,.7,1,0/236,101,0/EC6500}
\definecolor{tud9b}{cmyk/RGB/HTML}{0,1,.9,0/230,0,26/E6001A}
\definecolor{tud10b}{cmyk/RGB/HTML}{.4,1,0,0/166,0,132/A60084}
\definecolor{tud11b}{cmyk/RGB/HTML}{.7,1,0,0/114,16,133/721085}
\definecolor{tud1c}{cmyk/RGB/HTML}{1,.7,.2,0/0,78,138/004E8A}
\definecolor{tud2c}{cmyk/RGB/HTML}{1,.5,.2,0/0,104,157/00689D}
\definecolor{tud3c}{cmyk/RGB/HTML}{1,.2,.6,0/0,136,119/008877}
\definecolor{tud4c}{cmyk/RGB/HTML}{.6,.1,1,0/127,171,22/7FAB16}
\definecolor{tud5c}{cmyk/RGB/HTML}{.4,.1,1,0/177,189,0/B1BD00}
\definecolor{tud6c}{cmyk/RGB/HTML}{.2,.3,1,0/215,172,0/D7AC00}
\definecolor{tud7c}{cmyk/RGB/HTML}{.2,.5,1,0/210,135,0/D28700}
\definecolor{tud8c}{cmyk/RGB/HTML}{.2,.8,1,0/204,76,3/CC4C03}
\definecolor{tud9c}{cmyk/RGB/HTML}{.3,1,.9,0/185,15,34/B90F22}
\definecolor{tud10c}{cmyk/RGB/HTML}{.5,1,.3,0/149,17,105/951169}
\definecolor{tud11c}{cmyk/RGB/HTML}{.8,1,.2,0/97,28,115/611C73}
\definecolor{tud1d}{cmyk/RGB/HTML}{1,.9,.3,0/36,53,114/243572}
\definecolor{tud2d}{cmyk/RGB/HTML}{1,.7,.4,0/0,78,115/004E73}
\definecolor{tud3d}{cmyk/RGB/HTML}{1,.4,.7,0/0,113,94/00715E}
\definecolor{tud4d}{cmyk/RGB/HTML}{.7,.3,1,0/106,139,55/6A8B22}
\definecolor{tud5d}{cmyk/RGB/HTML}{.5,.2,1,0/153,166,4/99A604}
\definecolor{tud6d}{cmyk/RGB/HTML}{.4,.4,1,0/174,142,0/AE8E00}
\definecolor{tud7d}{cmyk/RGB/HTML}{.3,.6,1,0/190,111,0/BE6F00}
\definecolor{tud8d}{cmyk/RGB/HTML}{.4,.8,1,0/169,73,19/A94913}
\definecolor{tud9d}{cmyk/RGB/HTML}{.5,1,.9,0/156,28,38/961C26}
\definecolor{tud10d}{cmyk/RGB/HTML}{.7,1,.5,0/115,32,84/732054}
\definecolor{tud11d}{cmyk/RGB/HTML}{.9,1,.3,0/76,34,106/4C226A}

\definecolor{necgray}{HTML}{7d7d7d}
\definecolor{necblue}{HTML}{2c328f}
\definecolor{necbluedark}{HTML}{00285e}
\definecolor{necorange}{HTML}{eb6e00}
\definecolor{necyellow}{HTML}{fff484}
\definecolor{green}{HTML}{388a73}

\definecolor{qp51}{rgb}{0.5,0,0}
\definecolor{qp0}{rgb}{0,0,0.5}

\definecolor{tud3b6b1}{RGB}{63, 168, 97}
\definecolor{tud3b6b2}{RGB}{127, 180, 65}
\definecolor{tud3b6b3}{RGB}{190, 191, 32}

\definecolor{cvprblue}{rgb}{0.21,0.49,0.74}
\usepackage[pagebackref,breaklinks,colorlinks,citecolor=cvprblue]{hyperref}

\def\confYear{2024}

\title{A Perspective on Deep Vision Performance\\ with Standard Image and Video Codecs}

\newcommand{\authorstep}{\hspace{0.3cm}}
\newcommand{\affiliationstep}{\hspace{0.25cm}}
\newcommand{\imagesize}{0.1015\linewidth}

\author{Christoph Reich\textsuperscript{\normalfont{}1,2,3,5} 
\authorstep Oliver Hahn\textsuperscript{\normalfont{}1}
\authorstep Daniel Cremers\textsuperscript{\normalfont{}2,5}
\authorstep Stefan Roth\textsuperscript{\normalfont{}1,4}
\authorstep Biplob Debnath\textsuperscript{\normalfont{}3}
\and \textsuperscript{1}TU Darmstadt\affiliationstep \textsuperscript{2}TU Munich\affiliationstep \textsuperscript{3}NEC Laboratories America, Inc.\\ \textsuperscript{4}Hessian Center for AI (hessian.AI)\affiliationstep \textsuperscript{5}Munich Center for Machine Learning (MCML)
}

\renewcommand*{\paragraph}[1]{\smallskip\noindent\textbf{#1}\hspace{0.5em}}

\usepackage{fancyhdr}
\usepackage{setspace}

\fancypagestyle{firststyle}
{

    \fancyhf{}
    \lfoot{{\footnotesize\begin{spacing}{.5}\parbox{\linewidth}{\vspace{1.85em}
    To appear in \emph{Proceedings of the IEEE/CVF Conference on Computer Vision and Pattern Recognition}, 1\textsuperscript{st} Workshop on AI for Streaming @ CVPR, 2024.%
    \\\hrule\vspace{\baselineskip}
    \copyright~2024 IEEE. Personal use of this material is permitted. Permission from IEEE must be obtained for all other uses, in any current or future media, including reprinting/republishing this material for advertising or promotional purposes, creating new collective works, for resale or redistribution to servers or lists, or reuse of any copyrighted component of this work in other works. 
    }\end{spacing}}}
}

\begin{document}

{
\twocolumn[{
\renewcommand\twocolumn[1][]{#1}
\maketitle
\begin{center}
    \sffamily
    \setlength\tabcolsep{0.5pt}
    \renewcommand*{\arraystretch}{1.1}
    \vspace{-1.5em}
    \begin{tabular}{c c c c c c c | c c}
        \cellcolor{tud1a!20} & \multicolumn{6}{c|}{\cellcolor{tud1a!20}\large\textbf{JPEG coding}\vphantom{\Large I}} & 
        \multicolumn{2}{c}{\cellcolor{tud3a!20}\large\textbf{H.264 coding}\vphantom{\Large I}} \\[-1.25pt]
        
        \cellcolor{tud1a!20} & \multicolumn{2}{c}{\cellcolor{tud1a!20}Image classification} & 
        \multicolumn{2}{c}{\cellcolor{tud1a!20} Object detection} & 
        \multicolumn{2}{c|}{\cellcolor{tud1a!20}Semantic segmentation} & 
        \multicolumn{2}{c}{\cellcolor{tud3a!20}Optical flow estimation}\\[-1.5pt]

        \cellcolor{tud1a!20} \rotatebox{90}{$\;\;\;\;\;\;$Input} & 
        \cellcolor{tud1a!20}\begin{tikzpicture} \node[] at (0, 0) {\includegraphics[height=\imagesize]{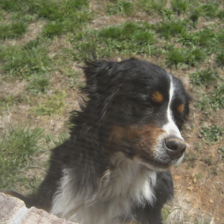}};\node[anchor=center, text width=1.25cm, align=center, color=white, fill=black,opacity=.4,text opacity=1] at (0, 0.675) {No coding};\end{tikzpicture} &
        \cellcolor{tud1a!20}\begin{tikzpicture} \node[] at (0, 0) {\includegraphics[height=\imagesize]{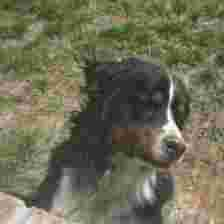}};\node[anchor=center, text width=1cm, align=center, color=white, fill=black,opacity=.4,text opacity=1] at (0, 0.675) {Coded\vphantom{g}};\end{tikzpicture} & 
        \cellcolor{tud1a!20}\begin{tikzpicture} \node[] at (0, 0) {\includegraphics[height=\imagesize]{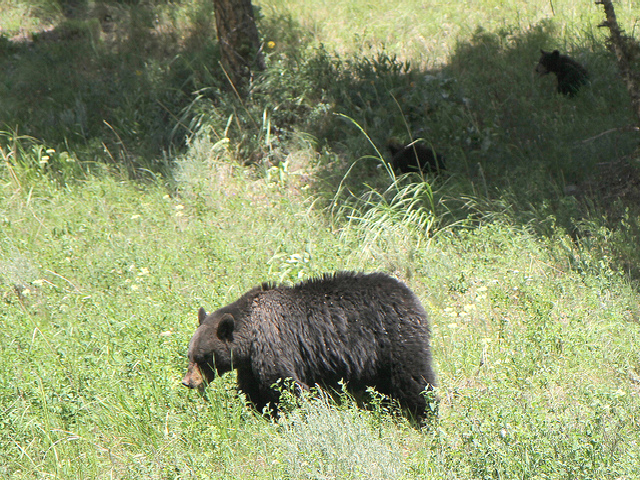}};\node[anchor=center, text width=1.25cm, align=center, color=white, fill=black,opacity=.4,text opacity=1] at (0, 0.675) {No coding};\end{tikzpicture} & 
        \cellcolor{tud1a!20}\begin{tikzpicture} \node[] at (0, 0) {\includegraphics[height=\imagesize]{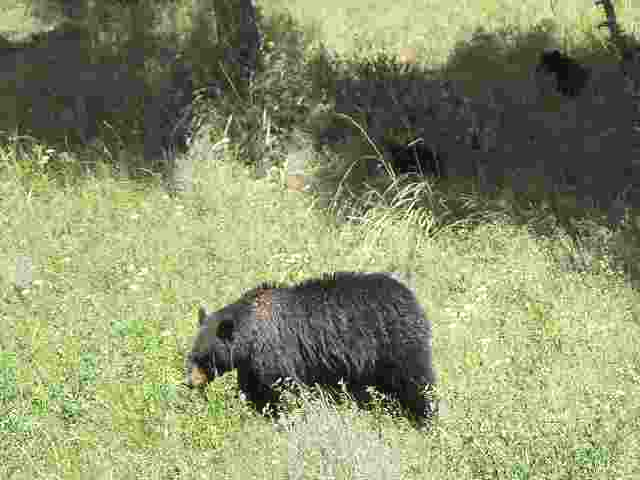}};\node[anchor=center, text width=1cm, align=center, color=white, fill=black,opacity=.4,text opacity=1] at (0, 0.675) {Coded\vphantom{g}};\end{tikzpicture} & 
        \cellcolor{tud1a!20}\begin{tikzpicture} \node[] at (0, 0) {\includegraphics[height=\imagesize]{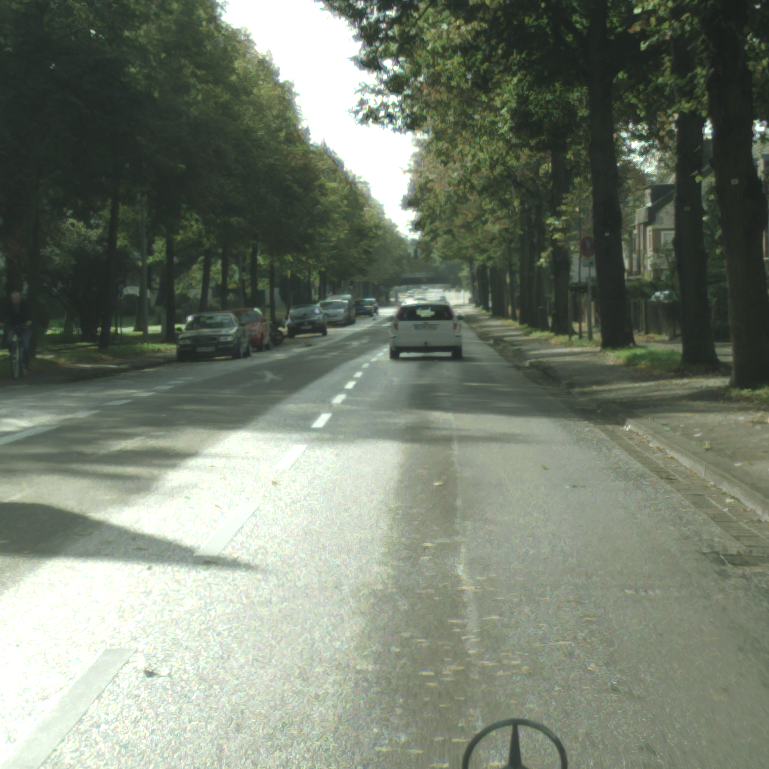}};\node[anchor=center, text width=1.25cm, align=center, color=white, fill=black,opacity=.4,text opacity=1] at (0, 0.675) {No coding};\end{tikzpicture} & 
        \cellcolor{tud1a!20}\begin{tikzpicture} \node[] at (0, 0) {\includegraphics[height=\imagesize]{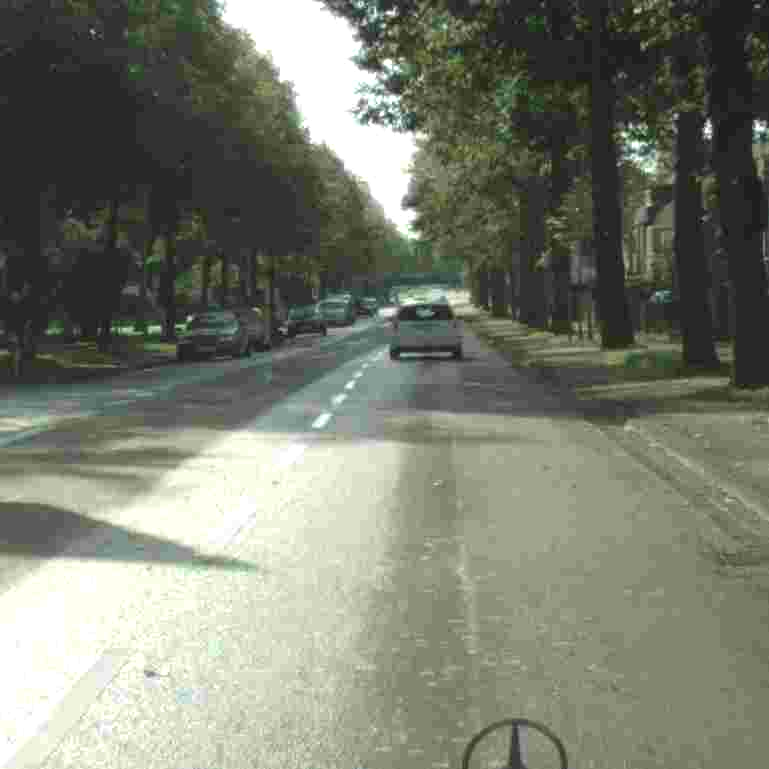}};\node[anchor=center, text width=1cm, align=center, color=white, fill=black,opacity=.4,text opacity=1] at (0, 0.675) {Coded\vphantom{g}};\end{tikzpicture} & 
        \cellcolor{tud3a!20}\begin{tikzpicture} \node[] at (0, 0) {\includegraphics[height=\imagesize]{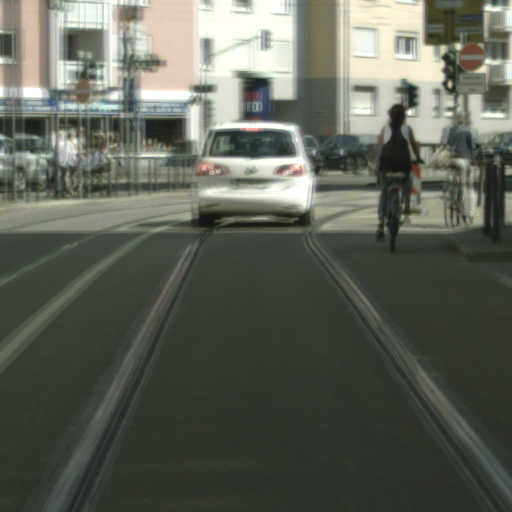}};\node[anchor=center, text width=1.25cm, align=center, color=white, fill=black,opacity=.4,text opacity=1] at (0, 0.675) {No coding};\end{tikzpicture} & 
        \cellcolor{tud3a!20}\cellcolor{tud3a!20}\begin{tikzpicture} \node[] at (0, 0) {\includegraphics[height=\imagesize]{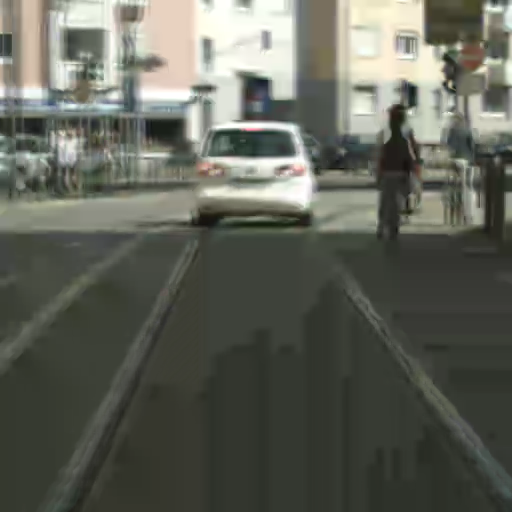}};\node[anchor=center, text width=1cm, align=center, color=white, fill=black,opacity=.4,text opacity=1] at (0, 0.675) {Coded\vphantom{g}};\end{tikzpicture} \\[-3pt]
        
        \cellcolor{tud1a!20} \rotatebox{90}{$\;\,$Prediction} & 

        \cellcolor{tud1a!20}\begin{tikzpicture}\node at (0, 0) {\fontsize{6}{7}\selectfont\begin{tabular}{c} \textbf{\underline{Top-3 pred.:}} \\ \textbf{Bernese dog} \\ Shetland dog \\ Border collie \end{tabular}};\node[color=tud1a!20] at (0, -0.75) {1904};\end{tikzpicture} &
        \cellcolor{tud1a!20}\begin{tikzpicture}\node at (0, 0) {\fontsize{6}{7}\selectfont\begin{tabular}{c} \textbf{\underline{Top-3 pred.:}} \\ \textbf{Wood rabbit}\\ Bluetick \\ Cardigan \end{tabular}};\node[color=tud1a!20] at (0, -0.75) {1904};\end{tikzpicture} & 
        \cellcolor{tud1a!20}\begin{tikzpicture} \node[] at (0, 0) {\includegraphics[height=\imagesize]{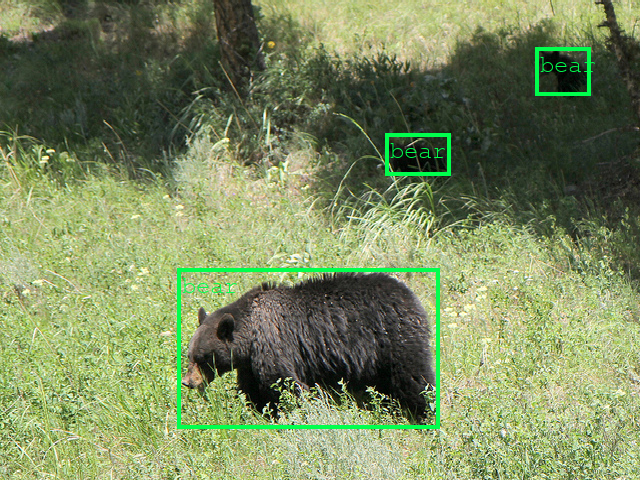}};\end{tikzpicture} & 
        \cellcolor{tud1a!20}\begin{tikzpicture} \node[] at (0, 0) {\includegraphics[height=\imagesize]{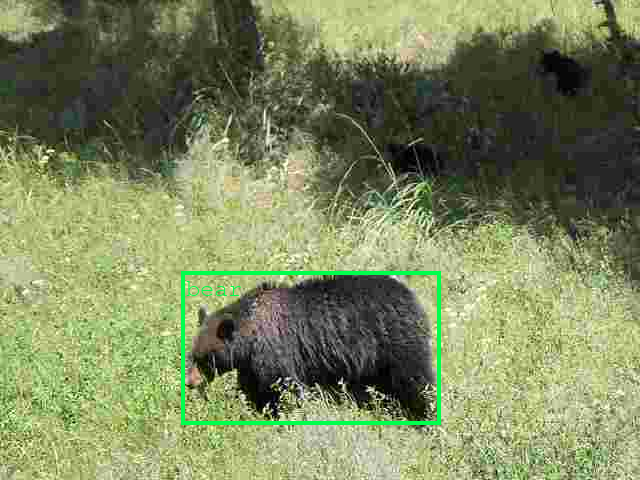}};\end{tikzpicture} & 
        \cellcolor{tud1a!20}\begin{tikzpicture} \node[] at (0, 0) {\includegraphics[height=\imagesize]{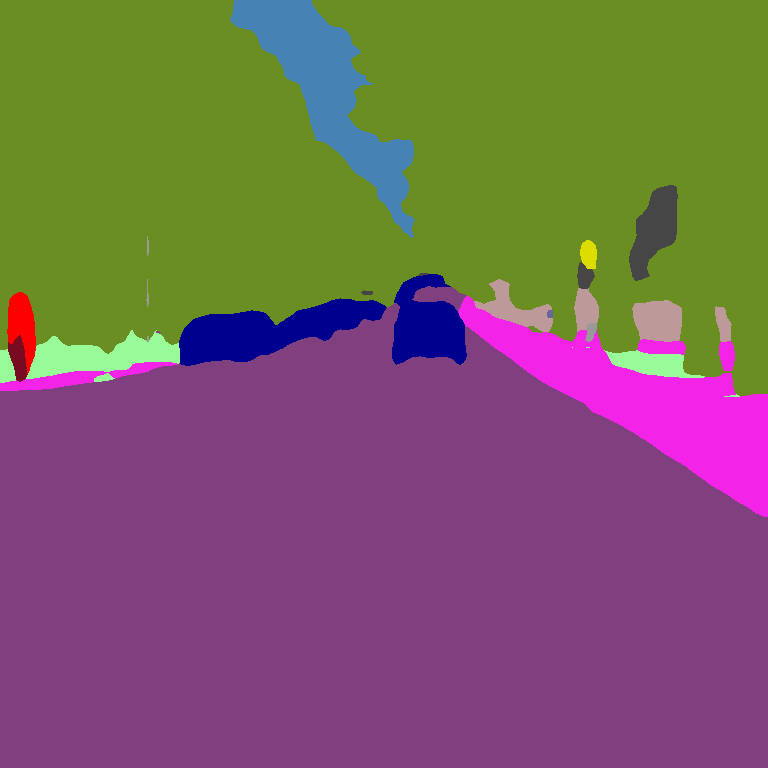}};\end{tikzpicture} & 
        \cellcolor{tud1a!20}\begin{tikzpicture} \node[] at (0, 0) {\includegraphics[height=\imagesize]{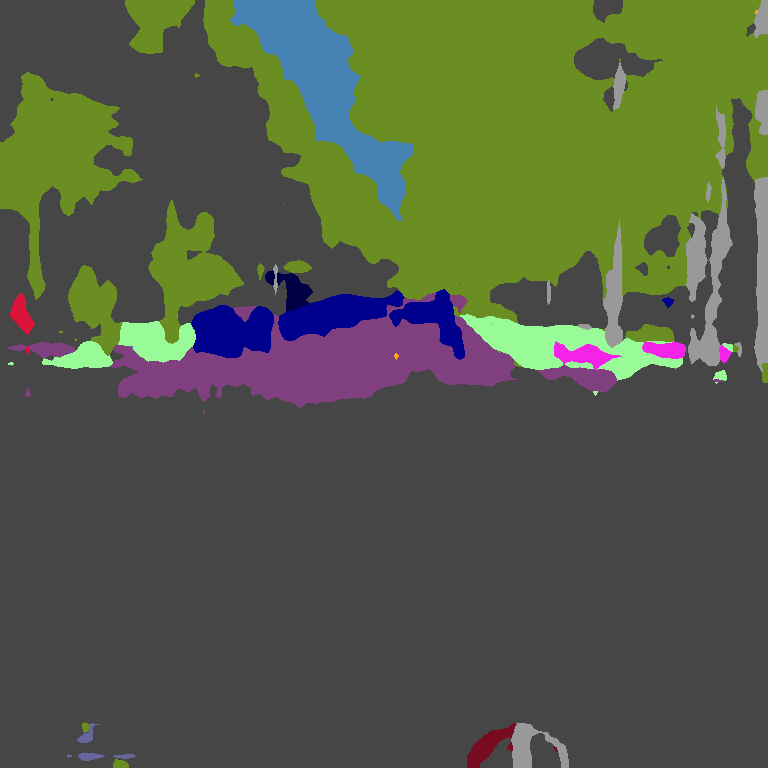}};\end{tikzpicture} & 
        \cellcolor{tud3a!20}\begin{tikzpicture} \node[] at (0, 0) {\includegraphics[height=\imagesize]{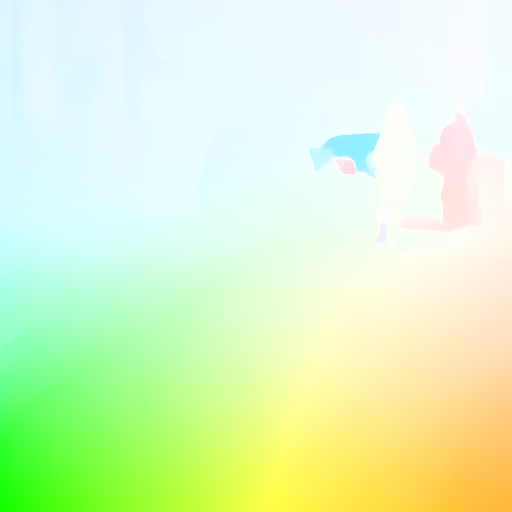}};\end{tikzpicture} & 
        \cellcolor{tud3a!20}\begin{tikzpicture} \node[] at (0, 0) {\includegraphics[height=\imagesize]{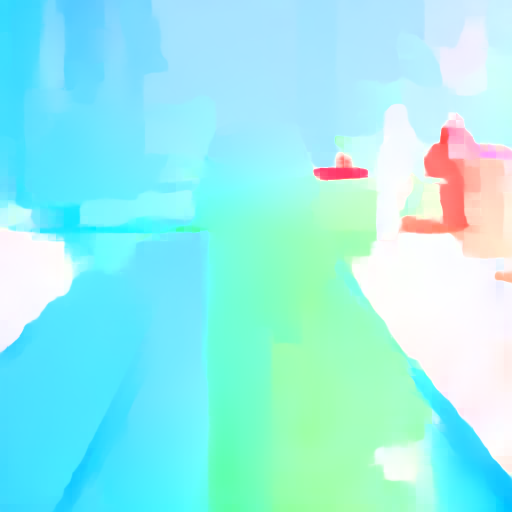}};\end{tikzpicture}
    \end{tabular}
    \vspace{-6pt}
    \captionof{figure}{\textbf{Deep vision performance on standard coded images/videos.} We demonstrate that using standard image and video coding can vastly deteriorate the accuracy of current deep vision models. We visualize the original \emph{(left)} and coded \emph{(right)} image/video and the respective models' prediction. For optical flow estimation, we overlay the first and second frame. Best viewed in color; zoom in for details.\label{fig:first}\\[-1pt]}
\end{center}
}]
}

\maketitle

\begin{abstract}
    Resource-constrained hardware, such as edge devices or cell phones, often rely on cloud servers to provide the required computational resources for inference in deep vision models. However, transferring image and video data from an edge or mobile device to a cloud server requires \emph{coding} to deal with network constraints. The use of standardized codecs, such as JPEG or H.264, is prevalent and required to ensure interoperability. This paper aims to examine the implications of employing standardized codecs within deep vision pipelines. We find that using JPEG and H.264 coding significantly deteriorates the accuracy across a broad range of vision tasks and models. For instance, strong compression rates reduce semantic segmentation accuracy by more than 80$\%$ in mIoU. In contrast to previous findings, our analysis extends beyond image and action classification to localization and dense prediction tasks, thus providing a more comprehensive perspective.
\end{abstract}
\thispagestyle{firststyle}

\section{Introduction}

Low-resource compute environments pose a particular challenge when analyzing images or videos. While efficient deep vision models for limited resource settings have been proposed, extensive image and video analysis is often facilitated using powerful cloud servers~\cite{Chen2020, Menghani2023}. In particular, for dense prediction tasks or in the presence of high-resolution imagery, edge or mobile devices transfer the respective data to cloud servers for inference~\cite{Du2022, Hu2023, Itsumi2022}. Data transfer is typically facilitated using wireless network connections.

{\begin{table*}[t!]
    \centering
    \caption{\textbf{Overview of the 23 evaluated models.} We provide both the source and unique identifier of each model used in our experiments.}
    \vspace{-5pt}
    {
\setlength\tabcolsep{0pt}
\renewcommand{\arraystretch}{0.8725}
\footnotesize
\begin{tabular}{
>{\raggedright\arraybackslash}p{0.28\textwidth}
>{\raggedright\arraybackslash}p{0.233\textwidth}
>{\raggedright\arraybackslash}p{0.487\textwidth}}
	\toprule
    $\,$Model & Source & Identifier \\
    \midrule
    \multicolumn{3}{c}{\cellcolor{tud1a!20}\textit{Semantic segmentation}} \\
    \cellcolor{tud1a!20}$\,$DeepLabV3 (ResNet-18)~\cite{Chen2017} & \cellcolor{tud1a!20}MMSegmentation~\cite{Mmseg2020} & \cellcolor{tud1a!20} \lstinline|deeplabv3_r18-d8_4xb2-80k_cityscapes-769x769|\\
    \cellcolor{tud1a!20}$\,$DeepLabV3 (ResNet-50)~\cite{Chen2017} & \cellcolor{tud1a!20}MMSegmentation~\cite{Mmseg2020} & \cellcolor{tud1a!20}\lstinline|deeplabv3_r50-d8_4xb2-40k_cityscapes-769x769| \\
    \cellcolor{tud1a!20}$\,$DeepLabV3 (ResNet-101)~\cite{Chen2017} & \cellcolor{tud1a!20}MMSegmentation~\cite{Mmseg2020} & \cellcolor{tud1a!20}\lstinline|deeplabv3_r101-d8_4xb2-40k_cityscapes-769x769| \\
    \cellcolor{tud1a!20}$\,$UPerNet (ResNet-50)~\cite{Xiao2018} & \cellcolor{tud1a!20}MMSegmentation~\cite{Mmseg2020} & \cellcolor{tud1a!20}\lstinline|upernet_r50_4xb2-40k_cityscapes-769x769| \\
    \cellcolor{tud1a!20}$\,$UPerNet (ResNet-101)~\cite{Xiao2018} & \cellcolor{tud1a!20}MMSegmentation~\cite{Mmseg2020} & \cellcolor{tud1a!20}\lstinline|upernet_r101_4xb2-40k_cityscapes-769x769| \\
    \midrule
    \multicolumn{3}{c}{\cellcolor{tud3a!20}\textit{Object detection}} \\
    \cellcolor{tud3a!20}$\,$DETR (ResNet-50)~\cite{Carion2020} & \cellcolor{tud3a!20}Official repository~\cite{Carion2020} & \cellcolor{tud3a!20} \lstinline|detr_resnet50| \\
    \cellcolor{tud3a!20}$\,$DETR (ResNet-101)~\cite{Carion2020} & \cellcolor{tud3a!20}Official repository~\cite{Carion2020} & \cellcolor{tud3a!20}\lstinline|detr_resnet101| \\
    \cellcolor{tud3a!20}$\,$DETR (ResNet-101-DC5)~\cite{Carion2020} & \cellcolor{tud3a!20}Official repository~\cite{Carion2020} & \cellcolor{tud3a!20}\lstinline|detr_resnet101_dc5| \\
    \cellcolor{tud3a!20}$\,$Faster R-CNN (ResNet-18)~\cite{Ren2017} & \cellcolor{tud3a!20}TorchVision~\cite{Torchvision2016} & \cellcolor{tud3a!20} \lstinline|fasterrcnn_resnet50_fpn(weights=DEFAULT)|\\
    \cellcolor{tud3a!20}$\,$Faster R-CNN V2 (ResNet-18)~\cite{Li2021} & \cellcolor{tud3a!20}TorchVision~\cite{Torchvision2016} & \cellcolor{tud3a!20} \lstinline|fasterrcnn_resnet50_fpn_v2(weights=DEFAULT)|\\
    \cellcolor{tud3a!20}$\,$Faster R-CNN (MobileNetV3 L)~\cite{Ren2017, Howard2019} & \cellcolor{tud3a!20}TorchVision~\cite{Torchvision2016} & \cellcolor{tud3a!20} \lstinline|fasterrcnn_mobilenet_v3_large_fpn(weights=DEFAULT)|\\
    \midrule
    \multicolumn{3}{c}{\cellcolor{tud6a!20}\textit{Image classification}} \\
    \cellcolor{tud6a!20}$\,$ResNet-18~\cite{He2016} & \cellcolor{tud6a!20}timm~\cite{Wightman2019} & \cellcolor{tud6a!20}\lstinline|resnet50(pretrained=True)| \\
    \cellcolor{tud6a!20}$\,$ResNet-50~\cite{He2016} & \cellcolor{tud6a!20}timm~\cite{Wightman2019} & \cellcolor{tud6a!20}\lstinline|resnet101(pretrained=True)| \\
    \cellcolor{tud6a!20}$\,$ResNet-101~\cite{He2016} & \cellcolor{tud6a!20}timm~\cite{Wightman2019} & \cellcolor{tud6a!20}\lstinline|resnet152(pretrained=True)| \\
    \cellcolor{tud6a!20}$\,$ViT-T~\cite{Dosovitskiy2020} & \cellcolor{tud6a!20}timm~\cite{Wightman2019} & \cellcolor{tud6a!20}\lstinline|vit_tiny_patch16_224(pretrained=True)| \\
    \cellcolor{tud6a!20}$\,$ViT-S~\cite{Dosovitskiy2020} & \cellcolor{tud6a!20}timm~\cite{Wightman2019} & \cellcolor{tud6a!20}\lstinline|vit_small_patch16_224(pretrained=True)| \\
    \cellcolor{tud6a!20}$\,$ViT-B~\cite{Dosovitskiy2020} & \cellcolor{tud6a!20}timm~\cite{Wightman2019} & \cellcolor{tud6a!20}\lstinline|vit_base_patch16_224(pretrained=True)| \\
    \cellcolor{tud6a!20}$\,$Swin-T~\cite{Liu2021b} & \cellcolor{tud6a!20}timm~\cite{Wightman2019} & \cellcolor{tud6a!20}\lstinline|swin_tiny_patch4_window7_224(pretrained=True)| \\
    \cellcolor{tud6a!20}$\,$Swin-S~\cite{Liu2021b} & \cellcolor{tud6a!20}timm~\cite{Wightman2019} & \cellcolor{tud6a!20}\lstinline|swin_small_patch4_window7_224(pretrained=True)| \\
    \cellcolor{tud6a!20}$\,$Swin-B~\cite{Liu2021b} & \cellcolor{tud6a!20}timm~\cite{Wightman2019} & \cellcolor{tud6a!20}\lstinline|swin_base_patch4_window7_224(pretrained=True)| \\
    \midrule
    \multicolumn{3}{c}{\cellcolor{tud11a!20}\textit{Optical flow estimation}} \\
    \cellcolor{tud11a!20}$\,$RAFT large~\cite{Teed2020} & \cellcolor{tud11a!20}TorchVision~\cite{Torchvision2016} & \cellcolor{tud11a!20}\lstinline|raft_large(weights=Raft_Large_Weights.C_T_V2)| \\
    \cellcolor{tud11a!20}$\,$RAFT small~\cite{Teed2020} & \cellcolor{tud11a!20}TorchVision~\cite{Torchvision2016} & \cellcolor{tud11a!20}\lstinline|raft_small(weights=Raft_Small_Weights.C_T_V2)| \\
    \cellcolor{tud11a!20}$\,$SMURF (RAFT large)~\cite{Teed2020} & \cellcolor{tud11a!20}Off. repo.~\cite{Stone2021} \& TorchVision~\cite{Torchvision2016} & \cellcolor{tud11a!20}\lstinline|raft_large(weights=Weights_From_Off_Repo)| \\
	\bottomrule
\end{tabular}}

    \label{tab:model_zoo}
\end{table*}}

Transferring image and video data over a network requires the use of coding to cope with network constraints, such as bandwidth limitations and fluctuations, to prevent data corruption~\cite{Itsumi2022}. To ensure low cost as well as interoperability, standardized codecs are the \emph{de facto} standard in real-world image and video processing pipelines~\cite{Hudson2018, Lederer2019}. However, computationally efficient standard codecs, such as JPEG~\cite{Wallace1992} or H.264~\cite{Richardson2004, Wiegand2003}, utilize lossy compression, leading to image/video distortions (\cf \cref{fig:first}).

While standard codecs had been developed prior to the deep learning era, targeting perceptual quality for humans~\cite{Wiegand2003}, they were na\"ively incorporated into deep vision pipelines~\cite{Du2022}. In this paper, we analyze the effect of standard image and video codecs on the predictive accuracy of common downstream deep vision models. While recent work demonstrated the effect of standard-coded image and video data on the task of action recognition~\cite{Otani2022} and image classification~\cite{Hendrycks2018, Shin2017}, we provide results on a wide range of different computer vision tasks, namely image classification, object detection, optical flow estimation, and semantic segmentation for a variety of approaches (\cf \cref{fig:first} \& \cref{tab:model_zoo}). We observe that all models tested significantly suffer from coding at inference time. In particular, standard coding leads to a significant decrease in model performance, especially for localization and dense prediction tasks (\cf \cref{fig:first}). When utilizing vast compression, required for limited bandwidth availability, the accuracy of deep vision methods completely breaks down (\cf \cref{fig:first}). For instance, when applying strong JPEG compression, the (absolute) ImageNet-1k~\cite{Russakovsky2015} classification accuracy of a Swin-B~\cite{Liu2021b} Transformer drops from about 83$\%$ to below 20$\%$.

We also discuss the limitations of recent approaches, such as optimizing standard codecs \wrt deep vision models~\cite{Du2022, Luo2021, Mandhane2022}. Based on the presented results and recent work, we close with a discussion of the challenges and possible solutions to overcome the vast deterioration of the predictive performance of deep vision models when employing standardized coding. With this, we hope to facilitate the development of novel approaches towards optimized coding for deep vision models within the scope of standardization.

\section{Experiments}

We aim to analyze the impact of standard image and video coding on the predictive performance of current deep vision models. Using the original model weights of the respective methods, we experiment with coded data during inference. Despite the availability of more advanced coding approaches, we use JPEG and H.264 coding as these are still the most commonly used codecs in real-world image/video processing pipelines~\cite{Hudson2018, Lederer2019}.

{\begin{figure*}[t]
    \centering
    \begin{tikzpicture}[every node/.style={font=\small}]
	\begin{groupplot}[
        group style={
            group name=ss,
            group size=2 by 1,
            ylabels at=edge left,
            horizontal sep=25pt,
            vertical sep=12pt,
        },
        legend style={nodes={scale=0.5}},
        height=3.5cm,
        xlabel shift=-1.5pt,
        width=0.5255\textwidth,
        grid=both,
        xtick pos=bottom,
        ytick pos=left,
        grid style={line width=.1pt, draw=white, dash pattern=on 1pt off 1pt},
        major grid style={line width=.2pt,draw=gray!50},
        minor tick num=1,
        xmin=0,
        xmax=100,
        ylabel shift=-1.5pt,
        xtick={
            1, 20, 40, 60, 80, 99
        },
        xticklabels={
            1, 20, 40, 60, 80, 99
        },
        ticklabel style = {font=\footnotesize},
        ]
        
        \nextgroupplot[
        title=DeepLabV3,
        title style={yshift=-6pt,},
        ylabel=mIoU (\%) $\uparrow$,
        xlabel=JPEG quality,
        ymin=0,
        ymax=100,
        ytick={
            0, 20, 40, 60, 80, 100
        },
        yticklabels={
            0, 20, 40, 60, 80, 100
        },
        ]
        \addplot[color=tud6b, mark=diamond*, mark size=0.9pt] table[x=x,y=miou] {jpeg_ss_deeplabv3_resnet50.dat};\label{pgfplots:ss_resnet50_bb};
        \addplot[color=tud1a, mark=*, mark size=0.9pt] table[x=x,y=miou] {jpeg_ss_deeplabv3_resnet18.dat};\label{pgfplots:ss_resnet18_bb};
        \addplot[color=tud8b, mark=square*, mark size=0.9pt] table[x=x,y=miou] {jpeg_ss_deeplabv3_resnet101.dat};\label{pgfplots:ss_resnet101_bb};
        
        \nextgroupplot[
        title=UPerNet\phantom{g},
        title style={yshift=-6pt,},
        xlabel=JPEG quality,
        ymin=0,
        ymax=100,
        ytick={
            0, 20, 40, 60, 80, 100
        },
        yticklabels={
            0, 20, 40, 60, 80, 100
        },
        ]
        \addplot[color=tud6b, mark=diamond*, mark size=0.9pt] table[x=x,y=miou] {jpeg_ss_upernet_resnet50.dat};
        \addplot[color=tud8b, mark=square*, mark size=0.9pt] table[x=x,y=miou] {jpeg_ss_upernet_resnet101.dat};
        
	\end{groupplot}
    \node[draw,fill=white, inner sep=0.5pt, anchor=center] at (0.335\textwidth, 0.4) {\tiny
        \begin{tabular}{
        p{\widthof{ResNet-101 backbone}}p{0.6cm}}
        ResNet-18 backbone & \ref*{pgfplots:ss_resnet18_bb} \\
        ResNet-50 backbone & \ref*{pgfplots:ss_resnet50_bb} \\
        ResNet-101 backbone & \ref*{pgfplots:ss_resnet101_bb} \\
        \end{tabular}};
\end{tikzpicture}
    \vspace{-5pt}
    \caption{\textbf{Relative semantic segmentation accuracy on JPEG-coded Cityscapes (val) dataset.} The accuracy of all models vastly decreases as the compression rate increases (lower JPEG quality). ResNet-18 backbone in \colorindicator{blue}{tud1a}, ResNet-50 backbone in \colorindicator{yellow}{tud6b}, and ResNet-101 backbone in \colorindicator{orange}{tud8b}. Best viewed in color.}
    \label{fig:jpeg_semantic_segmentation}
\end{figure*}
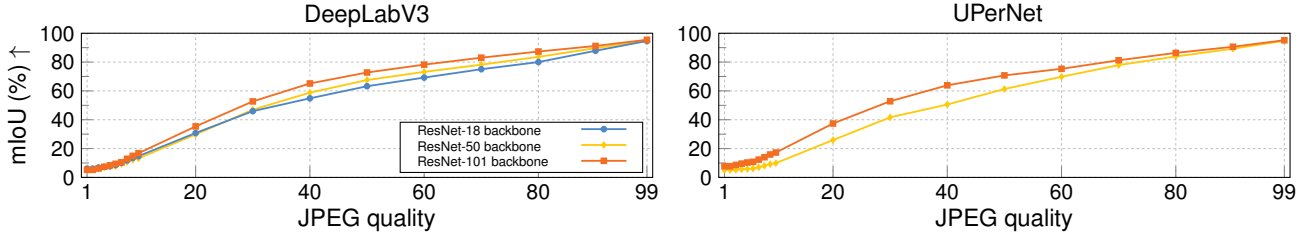}

\paragraph{Standard codecs.} Real-world image and video processing pipelines predominantly utilize standard codecs for transferring or storing image and video data to ensure interoperability. JPEG~\cite{Wallace1992} and H.264~\cite{Wiegand2003, Richardson2004} are among the most widely used standard codecs~\cite{Hudson2018, Lederer2019}. The popularity of both codecs in real-world applications is caused by their computational efficiency, acceptable rate-distortion trade-off, and widespread hardware and software support. Both codecs utilize quantization in the frequency domain to perform coding. By adjusting the quantization strength rate, the file size or required bandwidth can be traded off against the image distortion. While the rate-distortion trade-off of standard codecs has been studied extensively through the lens of Shannon's rate-distortion theory and by perceptual quality~\cite{Blau2019, Shannon1959}, we analyze the implications of standard codecs on the predictive performance of current deep vision models across a variety of downstream tasks.

\paragraph{Evaluation approach.} We aim to measure the effect of image and video coding on the accuracy of deep vision models, isolated from the models' respective baseline. To this end, we utilize the prediction on the original image as a reference, \ie, pseudo-ground truth and not the ground-truth label. If not stated differently, we perform inference using the original (non-coded) image or video and use the models' prediction as the pseudo label. In the case of classification tasks, \eg, semantic segmentation, we generate the pseudo labels by taking the $\arg\max$ of the models' output. We then measure the difference of the predictions, obtained on coded images or videos, \wrt the pseudo labels, resulting in a relative score on the respective downstream metric. This leads to a normalized and easy-to-compare measurement which reflects the amount of downstream performance in terms of accuracy lost due to coding. For all image classification methods, we also provide absolute scores \wrt the ground truth labels on ImageNet-1k~\cite{Russakovsky2015}, demonstrating the validity of our pseudo-label-based evaluation approach.

Depending on the vision task, we utilize task-specific metrics. For image classification, we report both the top-1 and top-5 accuracy (Acc). The mean average precision (mAP) is used to evaluate object detection. We also report mAP values using fixed intersection-over-union thresholds of 50$\%$ and 75$\%$. To analyze semantic segmentation predictions, we use the mean intersection-over-union~(mIoU) evaluation metric, computing the class-wise IoU before averaging over all classes. Finally, to evaluate optical flow estimation, we employ the average end-point error (EPE).

\paragraph{Experimental setup.} All utilized models are trained using their respective \emph{standard training protocols}. More precisely, we use publicly available pre-trained model weights of the respective methods (\cf \cref{tab:model_zoo}). For semantic segmentation, we use the DeepLabV3~\cite{Chen2017} and UPerNet~\cite{Xiao2018} models from MMSegmentation~\cite{Mmseg2020}. We vary the backbone between a ResNet-18 (only for DeepLabV3), 50, and 101~\cite{He2016}. Semantic segmentation experiments are performed on the Cityscapes \cite{Cordts2016} validation dataset. We follow the training resolution of the utilized models and evaluate using a resolution of 769$\times$769 pixels. The Detection Transformer (DETR)~\cite{Carion2020} model (w/ ResNet-50, 101 \& 101-DC5 backbones) from the official repository is used for object detection experiments. For the Faster R-CNN (V1 \& V2)~\cite{Ren2017, Li2021}, we utilize the checkpoints from TorchVision (w/ ResNet-50 and MobileNetV3 L~\cite{Howard2019} backbones). Object detection experiments are carried out on the Common Objects in Context (COCO)~\cite{Lin2014} dataset. We follow common evaluation and resize the images to a resolution of 480$\times$640 before performing coding. For the task of image classification, we employ different ResNet~\cite{He2016}, Vision Transformer~\cite{Dosovitskiy2020} (ViT), and Swin Transformer~\cite{Liu2021b} models from timm~\cite{Wightman2019}, trained on the ImageNet-1k \cite{Russakovsky2015} dataset. Following the timm evaluation protocol, we first resize the smaller image dimension to 256 pixels and keep the aspect ratio. Finally, we center-crop the images to a resolution of 224\textsuperscript{2} before performing coding. Optical flow estimation experiments are carried out on the supervised RAFT (large \& small)~\cite{Teed2020} from TorchVision~\cite{Torchvision2016} and the unsupervised SMURF~\cite{Stone2021} model (RAFT large architecture). We perform optical flow estimation on the Cityscapes validation sequences~\cite{Cordts2016} using a clip length of eight frames with a temporal stride of 3. We center-crop each video frame to a resolution of 512$\times$512 before performing video coding. In \cref{tab:model_zoo}, we provide a full overview of the utilized models.

{\begin{figure*}[t]
    \centering
    \input{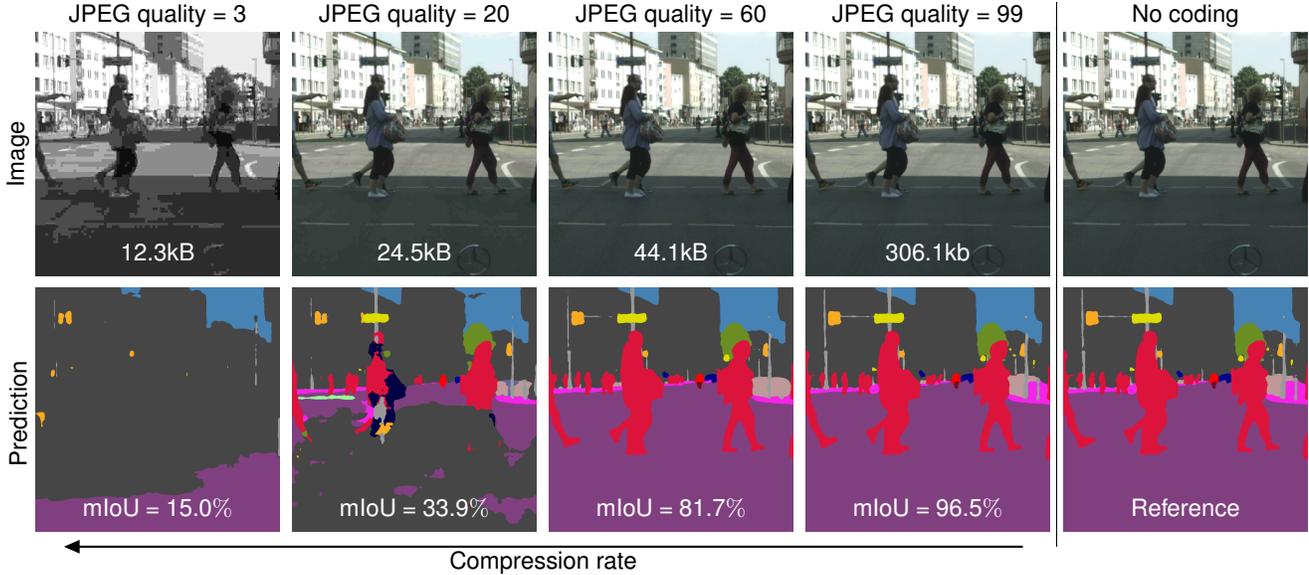}
    \vspace{-6pt}
    \caption{\textbf{Qualitative semantic segmentation example on a JPEG-coded Cityscapes (val) images.} As the compression rate increases (lower JPEG quality), the DeepLabV3 (w/ ResNet-18 backbone) model is not able to maintain its accuracy. For a JPEG quality of 3, the semantic segmentation accuracy completely breaks down. The mIoU relative to the uncoded baseline prediction is computed on a per-image basis. We report the JPEG file size for the coded images. Best viewed in color; zoom in for details.}
    \label{fig:jpeg_semantic_segmentation_plots}
\end{figure*}}{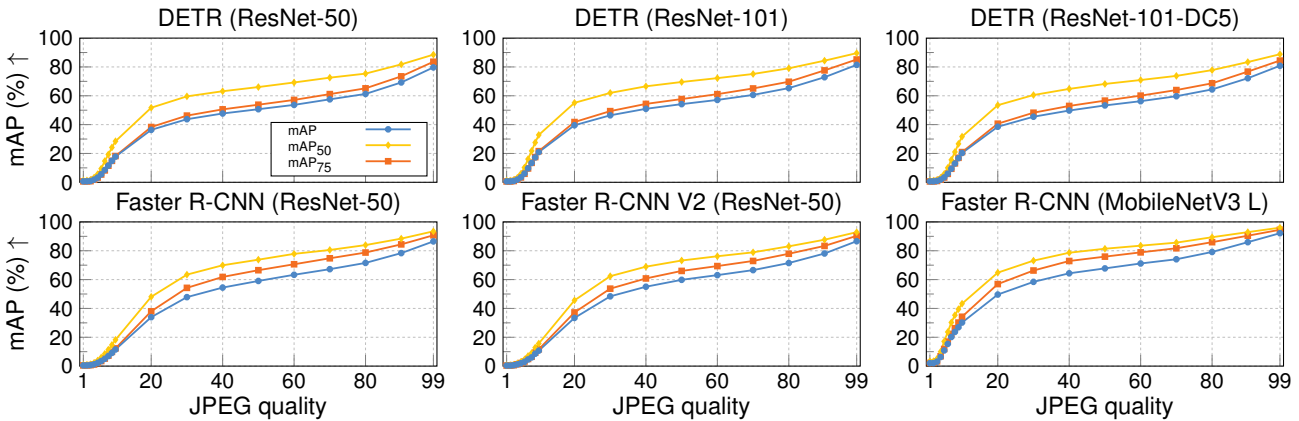
\begin{figure*}[t]
    \centering
    \begin{tikzpicture}[every node/.style={font=\small}]
	\begin{groupplot}[
        group style={
            group name=od,
            group size=3 by 3,
            ylabels at=edge left,
            horizontal sep=25pt,
            vertical sep=15pt,
        },
        legend style={nodes={scale=0.5}},
        height=3.5cm,
        xlabel shift=-1.5pt,
        width=0.3625\textwidth,
        grid=both,
        xtick pos=bottom,
        ytick pos=left,
        grid style={line width=.1pt, draw=white, dash pattern=on 1pt off 1pt},
        major grid style={line width=.2pt,draw=gray!50},
        minor tick num=1,
        xmin=0,
        xmax=100,
        ylabel shift=-1.5pt,
        xtick={
            1, 20, 40, 60, 80, 99
        },
        xticklabels={
            1, 20, 40, 60, 80, 99
        },
        ticklabel style = {font=\footnotesize},
        ]
        
        \nextgroupplot[
        title=DETR (ResNet-50),
        title style={yshift=-6pt,},
        ylabel=mAP (\%) $\uparrow$,
        ymin=0,
        ymax=100,
        ytick={0, 20, 40, 60, 80, 100},
        yticklabels={0, 20, 40, 60, 80, 100},
        xticklabels={,,},
        ]
    	\addplot[color=tud8b, mark=square*, mark size=0.9pt] table[x=x,y expr=100*\thisrow{map75}] {jpeg_od_detr_resnet50.dat}; \label{pgfplots:jpeg_resnet50_map75};
        \addplot[color=tud6b, mark=diamond*, mark size=0.9pt] table[x=x,y expr=100*\thisrow{map50}] {jpeg_od_detr_resnet50.dat}; \label{pgfplots:jpeg_resnet50_map50};
        \addplot[color=tud1a, mark=*, mark size=0.9pt] table[x=x,y expr=100*\thisrow{map}] {jpeg_od_detr_resnet50.dat}; \label{pgfplots:jpeg_resnet50_map};
        
        \nextgroupplot[
        title=DETR (ResNet-101),
        title style={yshift=-6pt,},
        ymin=0,
        ymax=100,
        ytick={0, 20, 40, 60, 80, 100},
        yticklabels={0, 20, 40, 60, 80, 100},
        xticklabels={,,},
        ]
    	\addplot[color=tud8b, mark=square*, mark size=0.9pt] table[x=x,y expr=100*\thisrow{map75}] {jpeg_od_detr_resnet101.dat};
        \addplot[color=tud6b, mark=diamond*, mark size=0.9pt] table[x=x,y expr=100*\thisrow{map50}] {jpeg_od_detr_resnet101.dat};
        \addplot[color=tud1a, mark=*, mark size=0.9pt] table[x=x,y expr=100*\thisrow{map}] {jpeg_od_detr_resnet101.dat};

        \nextgroupplot[
        title=DETR (ResNet-101-DC5),
        title style={yshift=-6pt,},
        ymin=0,
        ymax=100,
        ytick={0, 20, 40, 60, 80, 100},
        yticklabels={0, 20, 40, 60, 80, 100},
        xticklabels={,,},
        ]
    	\addplot[color=tud8b, mark=square*, mark size=0.9pt] table[x=x,y expr=100*\thisrow{map75}] {jpeg_od_detr_resnet101_dc5.dat};
        \addplot[color=tud6b, mark=diamond*, mark size=0.9pt] table[x=x,y expr=100*\thisrow{map50}] {jpeg_od_detr_resnet101_dc5.dat};
        \addplot[color=tud1a, mark=*, mark size=0.9pt] table[x=x,y expr=100*\thisrow{map}] {jpeg_od_detr_resnet101_dc5.dat};

        \nextgroupplot[
        title=Faster R-CNN (ResNet-50),
        title style={yshift=-6pt,},
        xlabel=JPEG quality,
        ylabel=mAP (\%) $\uparrow$,
        ymin=0,
        ymax=100,
        ytick={0, 20, 40, 60, 80, 100},
        yticklabels={0, 20, 40, 60, 80, 100},
        ]
    	\addplot[color=tud8b, mark=square*, mark size=0.9pt] table[x=x,y expr=100*\thisrow{map75}] {jpeg_od_faster_r_cnn.dat};
        \addplot[color=tud6b, mark=diamond*, mark size=0.9pt] table[x=x,y expr=100*\thisrow{map50}] {jpeg_od_faster_r_cnn.dat};
        \addplot[color=tud1a, mark=*, mark size=0.9pt] table[x=x,y expr=100*\thisrow{map}] {jpeg_od_faster_r_cnn.dat};

        \nextgroupplot[
        title=Faster R-CNN V2 (ResNet-50),
        title style={yshift=-6pt,},
        xlabel=JPEG quality,
        ymin=0,
        ymax=100,
        ytick={0, 20, 40, 60, 80, 100},
        yticklabels={0, 20, 40, 60, 80, 100},
        ]
    	\addplot[color=tud8b, mark=square*, mark size=0.9pt] table[x=x,y expr=100*\thisrow{map75}] {jpeg_od_faster_r_cnn_v2.dat};
        \addplot[color=tud6b, mark=diamond*, mark size=0.9pt] table[x=x,y expr=100*\thisrow{map50}] {jpeg_od_faster_r_cnn_v2.dat};
        \addplot[color=tud1a, mark=*, mark size=0.9pt] table[x=x,y expr=100*\thisrow{map}] {jpeg_od_faster_r_cnn_v2.dat};

        \nextgroupplot[
        title=Faster R-CNN (MobileNetV3 L),
        title style={yshift=-6pt,},
        xlabel=JPEG quality,
        ymin=0,
        ymax=100,
        ytick={0, 20, 40, 60, 80, 100},
        yticklabels={0, 20, 40, 60, 80, 100},
        ]
    	\addplot[color=tud8b, mark=square*, mark size=0.9pt] table[x=x,y expr=100*\thisrow{map75}] {jpeg_od_faster_r_cnn_mobilenet.dat};
        \addplot[color=tud6b, mark=diamond*, mark size=0.9pt] table[x=x,y expr=100*\thisrow{map50}] {jpeg_od_faster_r_cnn_mobilenet.dat};
        \addplot[color=tud1a, mark=*, mark size=0.9pt] table[x=x,y expr=100*\thisrow{map}] {jpeg_od_faster_r_cnn_mobilenet.dat};
        
	\end{groupplot}
	
	\node[draw,fill=white, inner sep=0.5pt, anchor=center] at (0.205\textwidth, 0.44) {\tiny
    \begin{tabular}{
    p{\widthof{mAP\textsubscript{50}}}p{0.6cm}}
    mAP\vphantom{(} & \ref*{pgfplots:jpeg_resnet50_map} \\
    mAP\textsubscript{50} & \ref*{pgfplots:jpeg_resnet50_map50} \\
    mAP\textsubscript{75\vphantom{(}} & \ref*{pgfplots:jpeg_resnet50_map75} \\
    \end{tabular}};
\end{tikzpicture}
    \vspace{-5pt}
    \caption{\textbf{Relative object detection accuracy on the JPEG-coded COCO (val) dataset.} The accuracy of all DETR and Faster R-CNN variants vastly deteriorates as the compression rate increases (lower JPEG quality). We report the mAP in \colorindicator{blue}{tud1a}, the mAP\textsubscript{50} in \colorindicator{yellow}{tud6b}, and the mAP\textsubscript{75} in \colorindicator{orange}{tud8b}. Best viewed in color.}
    \label{fig:jpeg_object_detection}
\end{figure*}}

\subsection{Predictive performance of deep vision models on JPEG-coded images}

We analyze the effect of JPEG coding on the predictive performance of deep vision models using different compression rates. JPEG offers to control the rate using the JPEG quality parameter. More specifically, the JPEG quality controls the quantization strength of the discrete cosine transform (DCT) coefficients and ranges from 1 to 99. A low JPEG quality corresponds to vast compression, leading to a small JPEG file size at the cost of image distortion. \emph{Vice versa}, a high JPEG quality leads to less compression.

{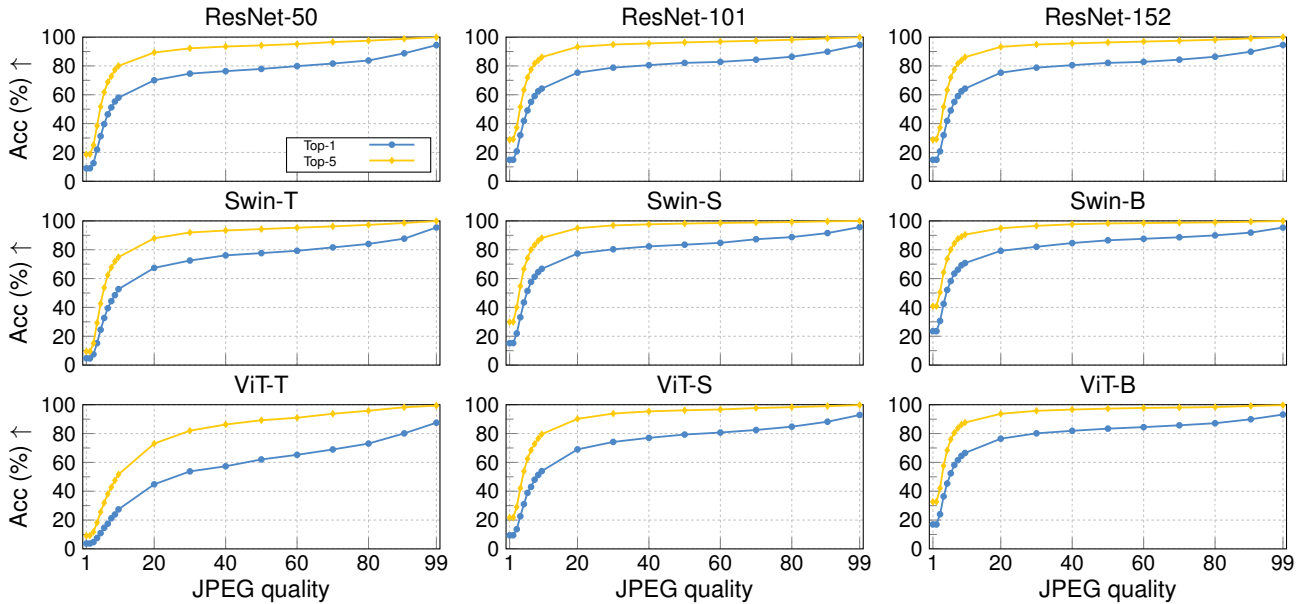
\begin{figure*}[t]
    \centering
    \begin{tikzpicture}[every node/.style={font=\small}]
	\begin{groupplot}[
        group style={
            group name=ic,
            group size=3 by 3,
            ylabels at=edge left,
            horizontal sep=25pt,
            vertical sep=15pt,
        },
        legend style={nodes={scale=0.5}},
        height=3.5cm,
        xlabel shift=-1.5pt,
        width=0.3625\textwidth,
        grid=both,
        xtick pos=bottom,
        ytick pos=left,
        grid style={line width=.1pt, draw=white, dash pattern=on 1pt off 1pt},
        major grid style={line width=.2pt,draw=gray!50},
        minor tick num=1,
        xmin=0,
        xmax=100,
        ylabel shift=-1.5pt,
        xtick={
            1, 20, 40, 60, 80, 99
        },
        xticklabels={
            1, 20, 40, 60, 80, 99
        },
        ticklabel style = {font=\footnotesize},
        ]
        
        \nextgroupplot[
        title=ResNet-50\vphantom{(},
        title style={yshift=-6pt,},
        ylabel=Acc (\%) $\uparrow$,
        xticklabels={,,},
        ytick={0, 20, 40, 60, 80, 100},
        yticklabels={0, 20, 40, 60, 80, 100},
        ymin=0,   
        ymax=100,
        ]
    	\addplot[color=tud1a, mark=*, mark size=0.9pt] table[x=x, y expr=100*\thisrow{acc1}] {jpeg_ic_resnet50.dat};
    	\addplot[color=tud6b, mark=diamond*, mark size=0.9pt] table[x=x, y expr=100*\thisrow{acc5}] {jpeg_ic_resnet50.dat};
        
        \nextgroupplot[
        title=ResNet-101\vphantom{(},
        title style={yshift=-6pt,},
        xticklabels={,,},
        ytick={0, 20, 40, 60, 80, 100},
        yticklabels={0, 20, 40, 60, 80, 100},
        ymin=0,   
        ymax=100,
        ]
    	\addplot[color=tud1a, mark=*, mark size=0.9pt] table[x=x, y expr=100*\thisrow{acc1}] {jpeg_ic_resnet101.dat};
    	\addplot[color=tud6b, mark=diamond*, mark size=0.9pt] table[x=x, y expr=100*\thisrow{acc5}] {jpeg_ic_resnet101.dat};
        
        \nextgroupplot[
        title=ResNet-152\vphantom{(},
        title style={yshift=-6pt,},
        xticklabels={,,},
        ytick={0, 20, 40, 60, 80, 100},
        yticklabels={0, 20, 40, 60, 80, 100},
        ymin=0,   
        ymax=100,
        ]
    	\addplot[color=tud1a, mark=*, mark size=0.9pt] table[x=x, y expr=100*\thisrow{acc1}] {jpeg_ic_resnet101.dat};
    	\addplot[color=tud6b, mark=diamond*, mark size=0.9pt] table[x=x, y expr=100*\thisrow{acc5}] {jpeg_ic_resnet101.dat};

        \nextgroupplot[
        title=Swin-T\vphantom{(},
        title style={yshift=-6pt,},
        ylabel=Acc (\%) $\uparrow$,
        xticklabels={,,},
        ytick={0, 20, 40, 60, 80, 100},
        yticklabels={0, 20, 40, 60, 80, 100},
        ymin=0,   
        ymax=100,
        ]
    	\addplot[color=tud1a, mark=*, mark size=0.9pt] table[x=x, y expr=100*\thisrow{acc1}] {jpeg_ic_swin_t.dat};
    	\addplot[color=tud6b, mark=diamond*, mark size=0.9pt] table[x=x, y expr=100*\thisrow{acc5}] {jpeg_ic_swin_t.dat};

        \nextgroupplot[
        title=Swin-S\vphantom{(},
        title style={yshift=-6pt,},
        xticklabels={,,},
        ytick={0, 20, 40, 60, 80, 100},
        yticklabels={0, 20, 40, 60, 80, 100},
        ymin=0,   
        ymax=100,
        ]
    	\addplot[color=tud1a, mark=*, mark size=0.9pt] table[x=x, y expr=100*\thisrow{acc1}] {jpeg_ic_swin_s.dat};
    	\addplot[color=tud6b, mark=diamond*, mark size=0.9pt] table[x=x, y expr=100*\thisrow{acc5}] {jpeg_ic_swin_s.dat};

        \nextgroupplot[
        title=Swin-B\vphantom{(},
        title style={yshift=-6pt,},
        xticklabels={,,},
        ytick={0, 20, 40, 60, 80, 100},
        yticklabels={0, 20, 40, 60, 80, 100},
        ymin=0,   
        ymax=100,
        ]
    	\addplot[color=tud1a, mark=*, mark size=0.9pt] table[x=x, y expr=100*\thisrow{acc1}] {jpeg_ic_swin_b.dat};
    	\addplot[color=tud6b, mark=diamond*, mark size=0.9pt] table[x=x, y expr=100*\thisrow{acc5}] {jpeg_ic_swin_b.dat};
    	
        \nextgroupplot[
        title=ViT-T\vphantom{(},
        ylabel=Acc (\%) $\uparrow$,
        title style={yshift=-6pt,},
        ytick={0, 20, 40, 60, 80, 100},
        yticklabels={0, 20, 40, 60, 80, 100},
        ymin=0,   
        ymax=100,
        xlabel=JPEG quality,
        ]
        \addplot[color=tud1a, mark=*, mark size=0.9pt] table[x=x, y expr=100*\thisrow{acc1}] {jpeg_ic_vit_t.dat}; \label{pgfplots:jpeg_vit_t_acc1};
        \addplot[color=tud6b, mark=diamond*, mark size=0.9pt] table[x=x, y expr=100*\thisrow{acc5}] {jpeg_ic_vit_t.dat}; \label{pgfplots:jpeg_vit_t_acc5};
        
        \nextgroupplot[
        title=ViT-S\vphantom{(},
        title style={yshift=-6pt,},
        ytick={0, 20, 40, 60, 80, 100},
        yticklabels={0, 20, 40, 60, 80, 100},
        ymin=0,   
        ymax=100,
        xlabel=JPEG quality,
        ]
        \addplot[color=tud1a, mark=*, mark size=0.9pt] table[x=x, y expr=100*\thisrow{acc1}] {jpeg_ic_vit_s.dat};
        \addplot[color=tud6b, mark=diamond*, mark size=0.9pt] table[x=x, y expr=100*\thisrow{acc5}] {jpeg_ic_vit_s.dat};
        
        \nextgroupplot[
        title=ViT-B\vphantom{(},
        title style={yshift=-6pt,},
        ytick={0, 20, 40, 60, 80, 100},
        yticklabels={0, 20, 40, 60, 80, 100},
        ymin=0,   
        ymax=100,
        xlabel=JPEG quality,
        ]
        \addplot[color=tud1a, mark=*, mark size=0.9pt] table[x=x, y expr=100*\thisrow{acc1}] {jpeg_ic_vit_b.dat};
        \addplot[color=tud6b, mark=diamond*, mark size=0.9pt] table[x=x, y expr=100*\thisrow{acc5}] {jpeg_ic_vit_b.dat};
        
	\end{groupplot}
	
	\node[draw,fill=white, inner sep=0.5pt, anchor=center] at (0.21\textwidth, 0.35) {\tiny
    \begin{tabular}{
    p{\widthof{Top-1}}p{0.6cm}}
    Top-1 & \ref*{pgfplots:jpeg_vit_t_acc1} \\
    Top-5 & \ref*{pgfplots:jpeg_vit_t_acc5} \\
    \end{tabular}};
\end{tikzpicture}
    \vspace{-5pt}
    \caption{\textbf{Relative image classification accuracy on the JPEG-coded ImageNet-1k (val) dataset.} The relative accuracy (\ie, \wrt the pseudo labels) of all models vastly deteriorates as the compression rate increases (lower JPEG quality). We report the top-1 accuracy in \colorindicator{blue}{tud1a} and the top-5 accuracy in \colorindicator{yellow}{tud6b}. Best viewed in color.}
    \label{fig:jpeg_image_classification}
\end{figure*}}
{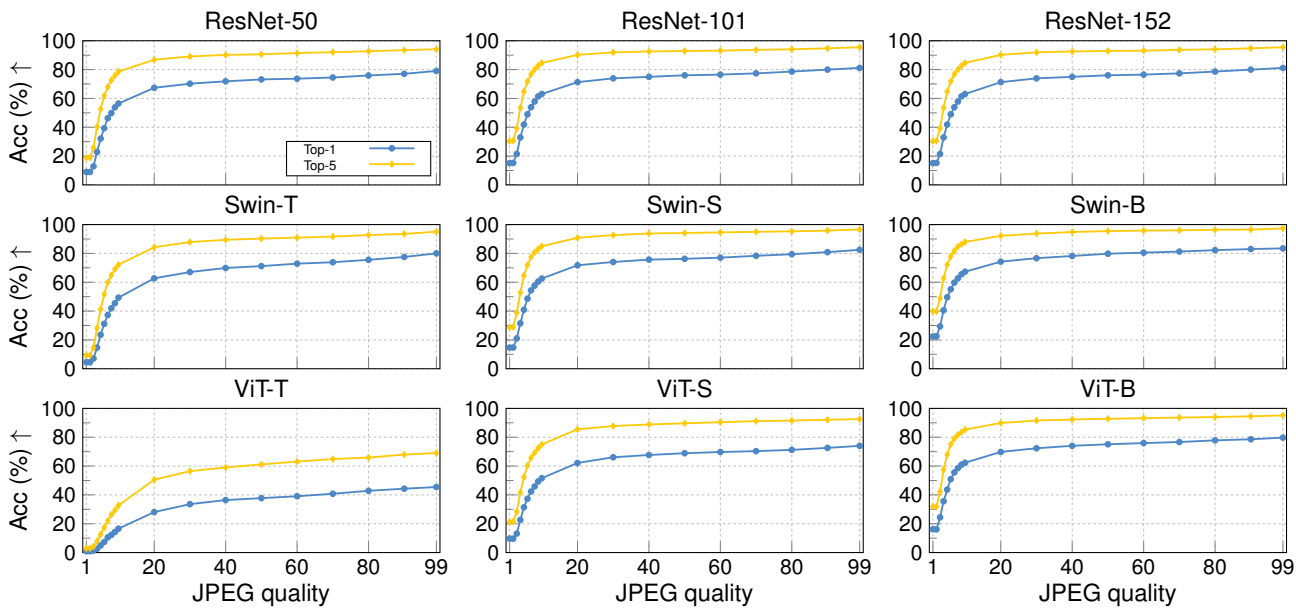
\begin{figure*}[t]
    \centering
    \begin{tikzpicture}[every node/.style={font=\small}]
	\begin{groupplot}[
        group style={
            group name=icgt,
            group size=3 by 3,
            ylabels at=edge left,
            horizontal sep=25pt,
            vertical sep=15pt,
        },
        legend style={nodes={scale=0.5}},
        height=3.5cm,
        xlabel shift=-1.5pt,
        width=0.3625\textwidth,
        grid=both,
        xtick pos=bottom,
        ytick pos=left,
        grid style={line width=.1pt, draw=white, dash pattern=on 1pt off 1pt},
        major grid style={line width=.2pt,draw=gray!50},
        minor tick num=1,
        xmin=0,
        xmax=100,
        ylabel shift=-1.5pt,
        xtick={
            1, 20, 40, 60, 80, 99
        },
        xticklabels={
            1, 20, 40, 60, 80, 99
        },
        ticklabel style = {font=\footnotesize},
        ]
        
        \nextgroupplot[
        title=ResNet-50\vphantom{(},
        title style={yshift=-6pt,},
        ylabel=Acc (\%) $\uparrow$,
        xticklabels={,,},
        ytick={0, 20, 40, 60, 80, 100},
        yticklabels={0, 20, 40, 60, 80, 100},
        ymin=0,   
        ymax=100,
        ]
    	\addplot[color=tud1a, mark=*, mark size=0.9pt] table[x=x, y expr=100*\thisrow{acc1}] {jpeg_ic_resnet50_gt.dat};
    	\addplot[color=tud6b, mark=diamond*, mark size=0.9pt] table[x=x, y expr=100*\thisrow{acc5}] {jpeg_ic_resnet50_gt.dat};
        
        \nextgroupplot[
        title=ResNet-101\vphantom{(},
        title style={yshift=-6pt,},
        xticklabels={,,},
        ytick={0, 20, 40, 60, 80, 100},
        yticklabels={0, 20, 40, 60, 80, 100},
        ymin=0,   
        ymax=100,
        ]
    	\addplot[color=tud1a, mark=*, mark size=0.9pt] table[x=x, y expr=100*\thisrow{acc1}] {jpeg_ic_resnet101_gt.dat};
    	\addplot[color=tud6b, mark=diamond*, mark size=0.9pt] table[x=x, y expr=100*\thisrow{acc5}] {jpeg_ic_resnet101_gt.dat};
        
        \nextgroupplot[
        title=ResNet-152\vphantom{(},
        title style={yshift=-6pt,},
        xticklabels={,,},
        ytick={0, 20, 40, 60, 80, 100},
        yticklabels={0, 20, 40, 60, 80, 100},
        ymin=0,   
        ymax=100,
        ]
    	\addplot[color=tud1a, mark=*, mark size=0.9pt] table[x=x, y expr=100*\thisrow{acc1}] {jpeg_ic_resnet101_gt.dat};
    	\addplot[color=tud6b, mark=diamond*, mark size=0.9pt] table[x=x, y expr=100*\thisrow{acc5}] {jpeg_ic_resnet101_gt.dat};

        \nextgroupplot[
        title=Swin-T\vphantom{(},
        title style={yshift=-6pt,},
        xticklabels={,,},
        ylabel=Acc (\%) $\uparrow$,
        ytick={0, 20, 40, 60, 80, 100},
        yticklabels={0, 20, 40, 60, 80, 100},
        ymin=0,   
        ymax=100,
        ]
    	\addplot[color=tud1a, mark=*, mark size=0.9pt] table[x=x, y expr=100*\thisrow{acc1}] {jpeg_ic_swin_t_gt.dat};
    	\addplot[color=tud6b, mark=diamond*, mark size=0.9pt] table[x=x, y expr=100*\thisrow{acc5}] {jpeg_ic_swin_t_gt.dat};

        \nextgroupplot[
        title=Swin-S\vphantom{(},
        title style={yshift=-6pt,},
        xticklabels={,,},
        ytick={0, 20, 40, 60, 80, 100},
        yticklabels={0, 20, 40, 60, 80, 100},
        ymin=0,   
        ymax=100,
        ]
    	\addplot[color=tud1a, mark=*, mark size=0.9pt] table[x=x, y expr=100*\thisrow{acc1}] {jpeg_ic_swin_s_gt.dat};
    	\addplot[color=tud6b, mark=diamond*, mark size=0.9pt] table[x=x, y expr=100*\thisrow{acc5}] {jpeg_ic_swin_s_gt.dat};

        \nextgroupplot[
        title=Swin-B\vphantom{(},
        title style={yshift=-6pt,},
        xticklabels={,,},
        ytick={0, 20, 40, 60, 80, 100},
        yticklabels={0, 20, 40, 60, 80, 100},
        ymin=0,   
        ymax=100,
        ]
    	\addplot[color=tud1a, mark=*, mark size=0.9pt] table[x=x, y expr=100*\thisrow{acc1}] {jpeg_ic_swin_b_gt.dat};
    	\addplot[color=tud6b, mark=diamond*, mark size=0.9pt] table[x=x, y expr=100*\thisrow{acc5}] {jpeg_ic_swin_b_gt.dat};
     
        \nextgroupplot[
        title=ViT-T\vphantom{(},
        ylabel=Acc (\%) $\uparrow$,
        title style={yshift=-6pt,},
        ytick={0, 20, 40, 60, 80, 100},
        yticklabels={0, 20, 40, 60, 80, 100},
        ymin=0,   
        ymax=100,
        xlabel=JPEG quality,
        ]
        \addplot[color=tud1a, mark=*, mark size=0.9pt] table[x=x, y expr=100*\thisrow{acc1}] {jpeg_ic_vit_t_gt.dat}; \label{pgfplots:jpeg_vit_t_acc1_gt};
        \addplot[color=tud6b, mark=diamond*, mark size=0.9pt] table[x=x, y expr=100*\thisrow{acc5}] {jpeg_ic_vit_t_gt.dat}; \label{pgfplots:jpeg_vit_t_acc5_gt};
        
        \nextgroupplot[
        title=ViT-S\vphantom{(},
        title style={yshift=-6pt,},
        ytick={0, 20, 40, 60, 80, 100},
        yticklabels={0, 20, 40, 60, 80, 100},
        ymin=0,   
        ymax=100,
        xlabel=JPEG quality,
        ]
        \addplot[color=tud1a, mark=*, mark size=0.9pt] table[x=x, y expr=100*\thisrow{acc1}] {jpeg_ic_vit_s_gt.dat};
        \addplot[color=tud6b, mark=diamond*, mark size=0.9pt] table[x=x, y expr=100*\thisrow{acc5}] {jpeg_ic_vit_s_gt.dat};
        
        \nextgroupplot[
        title=ViT-B\vphantom{(},
        title style={yshift=-6pt,},
        ytick={0, 20, 40, 60, 80, 100},
        yticklabels={0, 20, 40, 60, 80, 100},
        ymin=0,   
        ymax=100,
        xlabel=JPEG quality,
        ]
        \addplot[color=tud1a, mark=*, mark size=0.9pt] table[x=x, y expr=100*\thisrow{acc1}] {jpeg_ic_vit_b_gt.dat};
        \addplot[color=tud6b, mark=diamond*, mark size=0.9pt] table[x=x, y expr=100*\thisrow{acc5}] {jpeg_ic_vit_b_gt.dat};
        
	\end{groupplot}
	
	\node[draw,fill=white, inner sep=0.5pt, anchor=center] at (0.21\textwidth, 0.35) {\tiny
    \begin{tabular}{
    p{\widthof{Top-1}}p{0.6cm}}
    Top-1 & \ref*{pgfplots:jpeg_vit_t_acc1_gt} \\
    Top-5 & \ref*{pgfplots:jpeg_vit_t_acc5_gt} \\
    \end{tabular}};
\end{tikzpicture}
    \vspace{-5pt}
    \caption{\textbf{Absolute image classification accuracy on the JPEG-coded ImageNet-1k (val) dataset.} The absolute accuracy (\ie, \wrt the ground-truth labels) of all models vastly deteriorates as the compression rate increases (lower JPEG quality). We report the top-1 accuracy in \colorindicator{blue}{tud1a} and the top-5 accuracy in \colorindicator{yellow}{tud6b}. The general trend of the absolute scores aligns well with the relative scores reported in \cref{fig:jpeg_image_classification}. Best viewed in color.}
    \label{fig:jpeg_image_classification_gt}
\end{figure*}}

\paragraph{Semantic segmentation.} \cref{fig:jpeg_semantic_segmentation} demonstrates results of three different DeepLabV3 models~\cite{Chen2017} for JPEG-coded versions of Cityscapes~\cite{Cordts2016} validation images. For weak compression rates (JPEG quality $>$ 90), all models are able to largely maintain semantic segmentation accuracy. However, as the compression rate increases (JPEG quality reduces), the semantic segmentation accuracy constantly deteriorates. In the case of maximum compression (JPEG quality = 1), the semantic segmentation accuracy completely breaks down with an mIoU of below 10$\%$ relative to the baseline output on non-coded images. While all models struggle to maintain accuracy, a large backbone network (\eg, a ResNet-152) seems to offer slightly more robustness to JPEG coding especially for lower compression strengths.

We also analyze semantic segmentation results qualitatively. \cref{fig:jpeg_semantic_segmentation_plots} presents semantic segmentation predictions from a DeepLabV3 (w/ ResNet-18 backbone) on a non-coded image (right) and JPEG-coded images (left). While a JPEG quality of 99 (least compression) leads to minimal changes in the segmentation map, a JPEG quality of 60 already leads to some misclassification (\eg ``sidewalk'' and ``pole'' classified as ``fence''). A JPEG quality of 20 significantly deteriorates the semantic segmentation accuracy, resulting in the misclassification of large image sections as well as major classes such as ``road''. Some objects are still segmented accurately but most parts of the segmentation prediction suffer severely from JPEG coding. In contrast, the visual appearance of the coded image is rather similar to the original. A JPEG quality of 3 completely deteriorates the predictive performance of the semantic segmentation where the vast majority of pixels are wrongly classified.

\paragraph{Object detection.} We analyze the accuracy of three DETR~\cite{Carion2020} as well as three Faster R-CNN~\cite{Ren2017, Li2021} models on JPEG-coded COCO~\cite{Lin2014} validation images. The object detection performance of both models is presented in \cref{fig:jpeg_object_detection}. Similar to the task of semantic segmentation, we observe a strong drop in object detection performance of both models when the compression rate is increased. Even for weak compression rates (high JPEG quality values), the object detection performance is negatively impacted by JPEG coding. For strong compression rates (JPEG quality $<$ 5), the performance drops below an mAP score of 5$\%$ relative to the prediction on the uncoded images. Compared to semantic segmentation, the performance of the models remains slightly more stable for medium compression rates, which we attribute to the more spatially precise nature of the task. While only leading to a minor increase in robustness, a large backbone leads to a slight improvement on coded images. Notably, Faster R-CNN seems to suffer less in terms of relative object detection accuracy from medium compression strengths (JPEG quality within 30 to 80) than DETR when using a ResNet-50 backbone (\cf \cref{fig:jpeg_object_detection} (\emph{left column})).

\paragraph{Image classification.} In \cref{fig:jpeg_image_classification}, we present relative image classification results on JPEG-coded ImageNet-1k~\cite{Russakovsky2015} images. We tested three ResNet models~\cite{He2016}, three Swin models~\cite{Liu2021b} as well as three ViT models~\cite{Dosovitskiy2020}. The performance of all models decreases slightly for compression rates up to around 30. However, for medium JPEG qualities, the deterioration is not as severe as for semantic segmentation (\cf \cref{fig:jpeg_semantic_segmentation}) and object detection (\cf \cref{fig:jpeg_object_detection}). Comparing the model behaviors with those of semantic segmentation or object detection models, image classification performance remains largely stable for a wide range of compression rates, while the opposite occurs for the more granular tasks. The high sensitivity of dense prediction and localization tasks compared to image classification indicates that the global information is largely preserved in coded images, whereas coding leads to erroneous local representations.

We also report the absolute image classification performance in \cref{fig:jpeg_image_classification_gt}. While a weak compression (JPEG quality $>$ 95) maintains most of the models' accuracy, a vast compression (JPEG quality $<$ 20) leads to a severe drop in classification accuracy. A moderate JPEG quality of 80 leads to a 3.1$\%$ drop in image classification accuracy when using a ResNet-50, relative to no coding (standard validation accuracy 79.0$\%$)~\cite{Wightman2019, He2016}. A very weak compression rate (JPEG quality of 90) reduces the image classification accuracy still by 1.9$\%$. Considering the more incremental recent improvements in ImageNet classification accuracy~\cite{Wightman2019}, this is a significant deterioration. More importantly, we show that the models perform analogously in terms of absolute numbers when compared to our evaluation approach using the prediction of the non-coded images as pseudo labels. Yet, our evaluation approach leads to normalized results that are easier to compare and interpret between models.

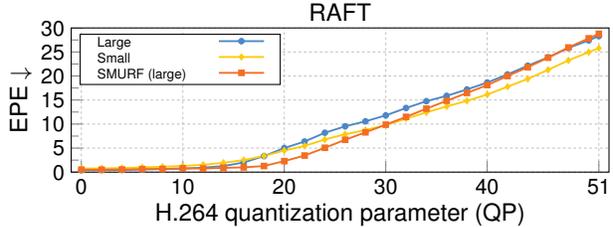
\begin{figure}[t!]
    \centering
    \begin{tikzpicture}[every node/.style={font=\small}]
	\begin{axis}[
        group style={
            group name=h264,
            group size=2 by 1,
            ylabels at=edge left,
            horizontal sep=37.5pt,
            vertical sep=12pt,
        },
        title=RAFT,
        title style={yshift=-6pt,},
        legend style={nodes={scale=0.5}},
        height=3.5cm,
        xlabel shift=-1.5pt,
        width=0.5\textwidth,
        grid=both,
        xtick pos=bottom,
        ytick pos=left,
        grid style={line width=.1pt, draw=white, dash pattern=on 1pt off 1pt},
        major grid style={line width=.2pt,draw=gray!50},
        minor tick num=1,
        xmin=-1,
        xmax=52,
        ylabel shift=-1.5pt,
        xtick={
            0, 10, 20, 30, 40, 51
        },
        xticklabels={
            0, 10, 20, 30, 40, 51
        },
        ticklabel style = {font=\footnotesize},
        ylabel=EPE $\downarrow$,
        xlabel=H.264 quantization parameter (QP),
        ymin=0,
        ymax=30,
        ytick={0, 5, 10, 15, 20, 25, 30},
        yticklabels={0, 5, 10, 15, 20, 25, 30},
        ]

    	\addplot[color=tud1a, mark=*, mark size=0.9pt] table[x=qp,y expr=\thisrow{mare}] {h264_of_raftlarge.dat};\label{pgfplots:h264_of_raftlarge};
        \addplot[color=tud6b, mark=diamond*, mark size=0.9pt] table[x=qp,y expr=\thisrow{mare}] {h264_of_raftsmall.dat};\label{pgfplots:h264_of_raftsmall};
        \addplot[color=tud8b, mark=square*, mark size=0.9pt] table[x=qp,y expr=\thisrow{mare}] {h264_of_smurf.dat};\label{pgfplots:h264_of_smurf};
        
	\end{axis}
    \node[draw,fill=white, inner sep=0.5pt, anchor=center] at (0.083\textwidth, 1.5) {\tiny
    \begin{tabular}{
    p{\widthof{SMURF (large)}}p{0.6cm}}
    Large & \ref*{pgfplots:h264_of_raftlarge} \\
    Small & \ref*{pgfplots:h264_of_raftsmall} \\
    SMURF (large) & \ref*{pgfplots:h264_of_smurf} \\
    \end{tabular}};
\end{tikzpicture}
    \vspace{-5pt}
    \caption{\textbf{Relative accuracy of optical flow on H.264-coded Cityscapes (val) clips.} Optical flow estimation of all three models vastly deteriorates as the compression rate increases (higher H.264 QP). We report the average EPE of RAFT large in \colorindicator{blue}{tud1a}, of RAFT small in \colorindicator{yellow}{tud6b}, and of SMURF (RAFT large unsupervised) in \colorindicator{orange}{tud8b}, relative to the respective baseline output. Best viewed in color.}
    \label{fig:h264_optical_flow}
\end{figure}

{\begin{figure*}[t]
    \centering
    \input{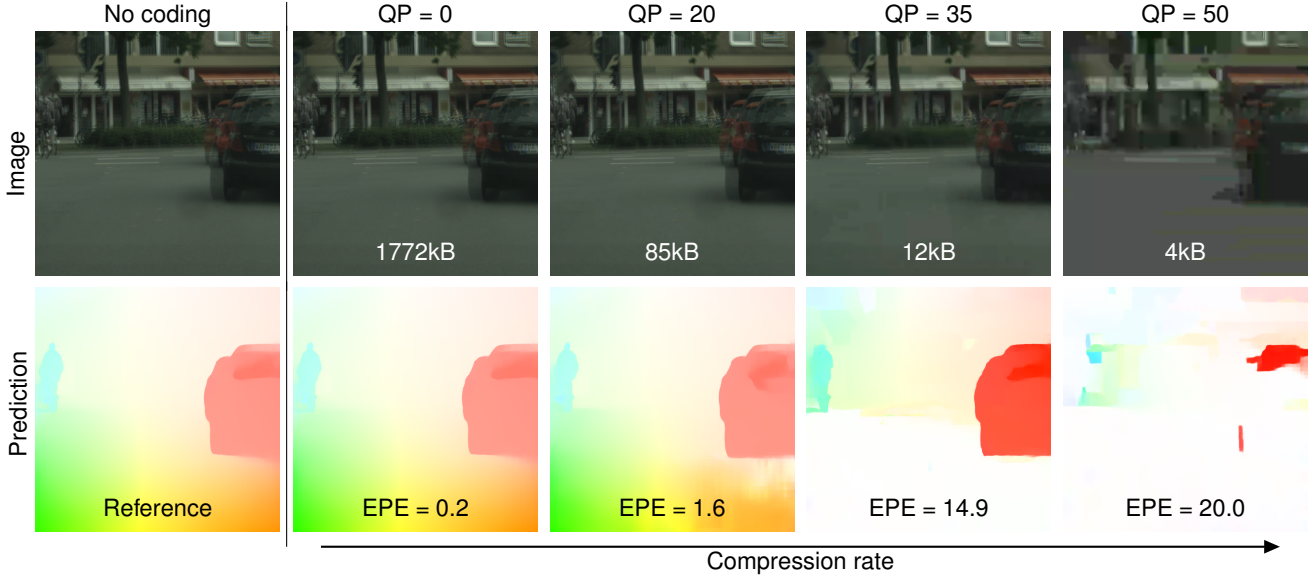}
    \vspace{-6pt}
    \caption{\textbf{Qualitative optical flow estimation example on H.264-coded Cityscapes (val) clips.} As the compression rate increases (higher QP), the optical flow estimation of RAFT large deteriorates. For a QP of 35, the predicted optical flow significantly deteriorates before completely breaking down for a QP of 50. We report the file size for the full 8-frame clip. We visualize the overlay of the first and second frame. Classical optical flow color encoding used as proposed by Baker \etal~\cite{Baker2011}. Best viewed in color; zoom in for details.}
    \label{fig:h264_optical_flow_plots}
\end{figure*}}

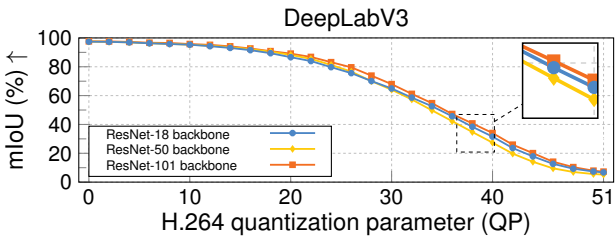
\begin{figure}[t!]
    \centering
    \begin{tikzpicture}[every node/.style={font=\small}, spy using outlines={black, line width=0.80mm, dashed, dash pattern=on 1.5pt off 1.5pt, magnification=2.0, size=1.0cm, connect spies, fill=white}]
	\begin{axis}[
        group style={
            group name=h264,
            group size=2 by 1,
            ylabels at=edge left,
            horizontal sep=37.5pt,
            vertical sep=12pt,
        },
        title=DeepLabV3,
        title style={yshift=-6pt,},
        legend style={nodes={scale=0.5}},
        height=3.5cm,
        xlabel shift=-1.5pt,
        width=0.49775\textwidth,
        grid=both,
        xtick pos=bottom,
        ytick pos=left,
        grid style={line width=.1pt, draw=white, dash pattern=on 1pt off 1pt},
        major grid style={line width=.2pt,draw=gray!50},
        minor tick num=1,
        xmin=-1,
        xmax=52,
        ylabel shift=-1.5pt,
        xtick={
            0, 10, 20, 30, 40, 51
        },
        xticklabels={
            0, 10, 20, 30, 40, 51
        },
        ticklabel style = {font=\footnotesize},
        ylabel=mIoU (\%) $\uparrow$,
        ymin=0,
        xlabel=H.264 quantization parameter (QP),
        ymax=100,
        ytick={0, 20, 40, 60, 80, 100},
        yticklabels={0, 20, 40, 60, 80, 100},
        ]
        \addplot[color=tud8b, mark=square*, mark size=0.9pt] table[x=qp,y expr=100*\thisrow{miou}] {h264_ss_deeplabv3resnet101.dat};\label{pgfplots:h264_ss_deeplabv3_resnet101}
        \addplot[color=tud6b, mark=diamond*, mark size=0.9pt] table[x=qp,y expr=100*\thisrow{miou}] {h264_ss_deeplabv3resnet50.dat};\label{pgfplots:h264_ss_deeplabv3_resnet50}
        \addplot[color=tud1a, mark=*, mark size=0.9pt] table[x=qp,y expr=100*\thisrow{miou}] {h264_ss_deeplabv3resnet18.dat};\label{pgfplots:h264_ss_deeplabv3_resnet18}

	\end{axis}
    \node[draw,fill=white, inner sep=0.5pt, anchor=center] at (0.1\textwidth, 0.41) {\tiny
    \begin{tabular}{
    p{\widthof{ResNet-101 backbone}}p{0.6cm}}
    ResNet-18 backbone & \ref*{pgfplots:h264_ss_deeplabv3_resnet18} \\
    ResNet-50 backbone & \ref*{pgfplots:h264_ss_deeplabv3_resnet50} \\
    ResNet-101 backbone & \ref*{pgfplots:h264_ss_deeplabv3_resnet101} \\
    \end{tabular}};
    \spy on (5.275, 0.65) in node [above, dash pattern=, line width=0.2mm, black, fill=white] at (6.4, 0.85);
\end{tikzpicture}
    \vspace{-5pt}
    \caption{\textbf{Relative accuracy of semantic segmentation on H.264-coded Cityscapes (val) clips.} The accuracy of all DeepLabV3 models vastly deteriorates as the compression rate increases (higher QP). We report the mIoU for different backbones (ResNet-18 in \colorindicatorn{blue}{tud1a}, 50 in \colorindicatorn{yellow}{tud6b} \& 101 in \colorindicatorn{orange}{tud8b}), relative to the baseline output. Best viewed in color.}
    \label{fig:h264_semantic_segmentation}
\end{figure}

\subsection{Predictive performance of deep vision models on H.264-coded videos}

We analyze the impact of H.264 video coding~\cite{Wiegand2003} on the predictive performance of deep vision models. To control the compression rate, H.264 offers support for different quantization parameters (QP). QP ranges from 0 to 51, where 0 corresponds to the least compression. In contrast, a QP of 51 leads to strong compression, resulting in significant video distortion at the benefit of a small rate. As we utilize validation clips of Cityscapes~\cite{Cordts2016} of eight frames, the group of images for H.264 coding is also set to eight. All other H.264 parameters are set to their respective default value~\cite{Tomar2006}. In particular, we use the FFmpeg command \texttt{ffmpeg -y -i frame\_\%10d.png -preset medium -g 8 -qp QP -flags +cgop+mv4 -mbtree 0 -aq-mode 0 -b\_strategy 0 -psy 0 video.mp4} and utilize different QP values.

\paragraph{Optical flow estimation.} In \cref{fig:h264_optical_flow}, we quantitatively evaluate the optical flow estimation performance of (supervised) RAFT (large \& small)~\cite{Teed2020} and (unsupervised) SMURF~\cite{Stone2021}. Note that the task of optical flow estimation relies on two consecutive frames, and thus also depends on the temporal quality of the coded clips. Optical flow estimation. In general, optical flow estimation performance for low QP values (below 15) remains fairly stable. As QP increases (QP $>$ 15) optical flow estimation deteriorates linearly. For QP $=$ 51 (strongest compression), the EPE of all models exceeds 25 \wrt the baseline output. While RAFT small offers an overall weaker optical flow estimation performance than RAFT large~\cite{Teed2020} (\wrt ground truth optical flow), the small RAFT variant tends to be more robust against H.264 coding than RAFT large, relative to the baseline output (pseudo label). Interestingly, while sharing the same architecture (RAFT large), unsupervised SMURF offers more robustness against H.264 than supervised RAFT large for QP values between 15 to 40. We attribute this slight improvement in robustness to the usage of strong data augmentation within the SMURF training approach~\cite{Stone2021}.

We also qualitatively analyze the optical flow estimation performance of (supervised) RAFT large~\cite{Teed2020} on an H.264 coded video clip in \cref{fig:h264_optical_flow_plots}. A QP of 0 as well as 20 only leads to a negligible change in the predicted optical flow map. Increasing QP to 35 leads to a notable deterioration of the predicted optical flow, especially in low-texture regions, such as the road. The apparent motion of object instances (\eg, car and cyclist) is still accurately estimated. For a strong compression rate (QP $=$ 50), also the estimated optical flow of moving object instances vastly deteriorates.

\paragraph{Semantic segmentation.} In \cref{fig:h264_semantic_segmentation}, we report quantitative semantic segmentation results of three different DeepLabV3~\cite{Chen2017} models on H.264-coded video clips. H.264 coding leads to a similar deterioration of segmentation accuracy as JPEG coding (\cf \cref{fig:jpeg_semantic_segmentation}). For low H.264 quantization parameters of up to approximately 15, the segmentation performance in mIoU is only affected slightly, whereas relative semantic segmentation accuracy decreases considerably in the subsequent medium and high compression range. For a QP above 40, the mIoU relative to the baseline output drops even below 40$\%$. For semantic segmentation on JPEG-coded images, we observed an increase in robustness against coding for larger backbones (\cf \cref{fig:jpeg_semantic_segmentation}). In the case of H.264 coded videos, all backbones (ResNet-18, 50 \& 101) suffer similarly from coding. In particular, DeepLabV3 with a ResNet-101 backbone offers better robustness against H.264 coding than the DeepLabV3 variant with a ResNet-18 backbone. The ResNet-50 DeepLabV3 variant suffers the most from H.264 coding.

\section{Discussion}

Relying on cloud servers for extensive processing of image and video data can, in principle, overcome bottlenecks from limited compute resources in mobile or edge devices. However, this approach requires the efficient and standardized transmission of image and video data. As we showed, using standardized codecs to facilitate efficient transmission can lead to a significantly reduced downstream deep vision performance. In our experiments, all methods suffered significantly in performance across a variety of vision tasks. Next, we will discuss different avenues to overcome the negative impact of standardized coding on deep vision models.

\subsection{Optimizing standard codecs}

Recent approaches have aimed to optimize the codec's parameters for deep vision models without breaking standardization. Luo~\etal~\cite{Luo2021} proposed optimized JPEG quantization tables for image classification using end-to-end learning. This approach, however, requires the typically non-differentiable image codec to be differentiable. While it is feasible to find differentiable surrogates of simple codecs, such as JPEG~\cite{Reich2024, Shin2017}, an increasing complexity of the image or video codec makes developing or learning a differentiable codec surrogate more challenging~\cite{Isik2023, Tian2021}. Du~\etal~\cite{Du2022} proposed an approach to optimize the allocation of the H.264 quantization parameters for vision models, which does not require a differentiable codec. Learning is performed on a saliency-based proxy task. However, learning on a proxy task can introduce unique challenges. The approach is also tailored to the task of object detection. A promising avenue for future research could be to build more powerful codec surrogate models to circumvent proxy task training. Alternatively, the use of reinforcement learning could overcome the need for a differentiable codec surrogate~\cite{Mandhane2022}. Additionally, offering support for rate control and multiple computer vision tasks in these approaches could facilitate a potentially widespread application of methods augmenting the parameters of standard codecs.

\subsection{Deep codecs for deep vision models}

Deep codecs aim to learn image and video coding using deep neural networks~\cite{Mishra2022, Zhang2023}. While reaching a better rate-distortion trade-off than standard codecs and offering support for custom objectives~\cite{Chamain2021, Choi2022, Chen2023, Le2021} (\eg, maintaining the accuracy of a downstream deep vision model), these approaches entail three major limitations so far. First, no standardization for deep codecs has emerged yet, severely limiting their application in real-world applications. Second, deep codecs only offer limited support for rate control. Finally, deep codecs are significantly less computationally efficient than standard codecs. This poses challenges, especially in low-resource environments. While offering great support for custom downstream objectives, optimizing a deep codec for multiple downstream vision tasks and models is non-trivial~\cite{Chen2023}. Recently, Chen~\etal~\cite{Chen2023} proposed a promising prompting-based approach to facilitate the support for multiple vision tasks and models. Overcoming these existing limitations and offering support for multiple vision tasks and models could potentially lead to the adoption of deep codecs in real-world deep learning pipelines.

\subsection{Data augmentation}

Using coding as a data augmentation strategy could enable models to be more robust to coding artifacts. As shown by Otani~\etal~\cite{Otani2022} and Reich~\etal~\cite{Reich2023}, coding-based data augmentation can introduce some degree of robustness, however, does not fully mitigate the deterioration in downstream accuracy as relevant information is still lost during coding. Additionally, training each downstream deep vision model from scratch using a custom augmentation pipeline based on the employed codec might be not feasible.

\subsection{Adapting deep vision models for coded data}

An alternative approach to overcome the deterioration in accuracy on coded images and videos is to augment the architecture of deep vision models. As demonstrated by Park \etal~\cite{Park2023} ViTs~\cite{Dosovitskiy2020} can consume JPEG DCT features instead of RGB images. This can introduce some degree of robustness to JPEG coding. However, similar to data augmentation, adapting all current deep architectures for custom representations of standard image and video codecs (\eg, JPEG DCT features or H.264 byte code) is non-trivial, requires considerable effort, and might be impractical.

\subsection{Dataset design and data pre-processing}

While beyond the scope of the paper and not considered in our experiments, some datasets already entail some degree of compression (\eg, ImageNet~\cite{Russakovsky2015} and Kinetics-400~\cite{Kay2017}). For instance, Kinetics-400 contains videos scraped from the internet compressed with various compression rates~\cite{Kay2017}. Additionally, pre-processing large video and image datasets often utilizes standard coding~\cite{Mmaction22020, Leclerc2023}. For example, MMAction2~\cite{Mmaction22020} typically extracts individual video frames from standard-coded videos using JPEG coding, resulting in frames that are coded multiple times. As we demonstrate standard coding can significantly impact downstream deep vision performance, thus, the careful consideration of coding during the design stage of datasets as well as during pre-processing is crucial.

\section{Conclusion}

In this paper, we analyzed the effect of standard image and video coding on four common vision tasks: image classification, object detection, semantic segmentation, as well as optical flow estimation. We tested a wide variety of current deep vision models (23 vision models in total) and observed that the predictive accuracy severely deteriorated with standard coding. Even a complete breakdown of downstream performance can arise for strong compression rates. Our analysis shows that image and video coding is particularly problematic for localization and dense prediction tasks. We discussed both challenges as well as possible avenues to mitigate the negative effects of (standardized) coding. We hope that our findings contribute to facilitating the development of improved standardized image and video codecs for deep vision models and shed light on the implications of deploying standardized codecs with deep vision models.

{\small
\paragraph{Acknowledgements.} This project was supported by the European Research Council (ERC) Advanced Grant SIMULACRON. This project has also received funding from the ERC under the European Union’s Horizon 2020 research and innovation programme (grant agreement No.\ 866008). In part, the project has been supported further by the State of Hesse through the cluster project “The Third Wave of Artificial Intelligence (3AI)“. Moreover, we thank NEC Laboratories America, Inc., for the support. Finally, we thank Deep Patel and Srimat Chakradhar for the insightful discussions.
}

{
    \small
    \bibliographystyle{ieeenat_fullname}
    \bibliography{main}

\begin{thebibliography}{50}
\providecommand{\natexlab}[1]{#1}
\providecommand{\url}[1]{\texttt{#1}}
\expandafter\ifx\csname urlstyle\endcsname\relax
  \providecommand{\doi}[1]{doi: #1}\else
  \providecommand{\doi}{doi: \begingroup \urlstyle{rm}\Url}\fi

\bibitem[Baker et~al.(2011)Baker, Scharstein, Lewis, Roth, Black, and
  Szeliski]{Baker2011}
Simon Baker, Daniel Scharstein, James~P Lewis, Stefan Roth, Michael~J Black,
  and Richard Szeliski.
\newblock A database and evaluation methodology for optical flow.
\newblock \emph{{IJCV}}, 92:\penalty0 1--31, 2011.

\bibitem[Blau and Michaeli(2019)]{Blau2019}
Yochai Blau and Tomer Michaeli.
\newblock Rethinking lossy compression: The rate-distortion-perception
  tradeoff.
\newblock In \emph{{ICML}}, pages 675--685, 2019.

\bibitem[Carion et~al.(2020)Carion, Massa, Synnaeve, Usunier, Kirillov, and
  Zagoruyko]{Carion2020}
Nicolas Carion, Francisco Massa, Gabriel Synnaeve, Nicolas Usunier, Alexander
  Kirillov, and Sergey Zagoruyko.
\newblock End-to-end object detection with transformers.
\newblock In \emph{{ECCV}}, pages 213--229, 2020.

\bibitem[Chamain et~al.(2021)Chamain, Racap{\'e}, B{\'e}gaint, Pushparaja, and
  Feltman]{Chamain2021}
Lahiru~D Chamain, Fabien Racap{\'e}, Jean B{\'e}gaint, Akshay Pushparaja, and
  Simon Feltman.
\newblock End-to-end optimized image compression for machines, a study.
\newblock In \emph{{DCC}}, pages 163--172, 2021.

\bibitem[Chen et~al.(2020)Chen, Zhang, Zhang, Dai, Yi, Zhang, and
  Zhang]{Chen2020}
Chunlei Chen, Peng Zhang, Huixiang Zhang, Jiangyan Dai, Yugen Yi, Huihui Zhang,
  and Yonghui Zhang.
\newblock Deep learning on computational-resource-limited platforms: A survey.
\newblock \emph{{Mob. Inf. Syst.}}, 2020:\penalty0 1--19, 2020.

\bibitem[Chen et~al.(2017)Chen, Papandreou, Schroff, and Adam]{Chen2017}
Liang-Chieh Chen, George Papandreou, Florian Schroff, and Hartwig Adam.
\newblock Rethinking atrous convolution for semantic image segmentation.
\newblock \emph{{arXiv:1706.05587 [cs.CV]}}, 2017.

\bibitem[Chen et~al.(2023)Chen, Weng, Kao, Chien, Chiu, and Peng]{Chen2023}
Yi-Hsin Chen, Ying-Chieh Weng, Chia-Hao Kao, Cheng Chien, Wei-Chen Chiu, and
  Wen-Hsiao Peng.
\newblock {TransTIC}: {T}ransferring transformer-based image compression from
  human perception to machine perception.
\newblock In \emph{{ICCV}}, pages 23297--23307, 2023.

\bibitem[Choi and Baji{\'c}(2022)]{Choi2022}
Hyomin Choi and Ivan~V Baji{\'c}.
\newblock Scalable image coding for humans and machines.
\newblock \emph{{IEEE Trans. Image Process.}}, 31:\penalty0 2739--2754, 2022.

\bibitem[Contributors(2020)]{Mmaction22020}
MMAction2 Contributors.
\newblock {OpenMMLab's} next generation video understanding toolbox and
  benchmark.
\newblock \url{https://github.com/open-mmlab/mmaction2}, 2020.

\bibitem[Cordts et~al.(2016)Cordts, Omran, Ramos, Rehfeld, Enzweiler, Benenson,
  Franke, Roth, and Schiele]{Cordts2016}
Marius Cordts, Mohamed Omran, Sebastian Ramos, Timo Rehfeld, Markus Enzweiler,
  Rodrigo Benenson, Uwe Franke, Stefan Roth, and Bernt Schiele.
\newblock The {Cityscapes} dataset for semantic urban scene understanding.
\newblock In \emph{{CVPR}}, pages 3213--3223, 2016.

\bibitem[Dosovitskiy et~al.(2020)Dosovitskiy, Beyer, Kolesnikov, Weissenborn,
  Zhai, Unterthiner, Dehghani, Minderer, Heigold, Gelly, Uszkoreit, and
  Houlsby]{Dosovitskiy2020}
Alexey Dosovitskiy, Lucas Beyer, Alexander Kolesnikov, Dirk Weissenborn,
  Xiaohua Zhai, Thomas Unterthiner, Mostafa Dehghani, Matthias Minderer, Georg
  Heigold, Sylvain Gelly, Jakob Uszkoreit, and Neil Houlsby.
\newblock An image is worth 16$\times$16 words: Transformers for image
  recognition at scale.
\newblock In \emph{{ICLR}}, 2020.

\bibitem[Du et~al.(2022)Du, Zhang, Arapin, Wang, Xia, and Jiang]{Du2022}
Kuntai Du, Qizheng Zhang, Anton Arapin, Haodong Wang, Zhengxu Xia, and Junchen
  Jiang.
\newblock {Acc\-MPEG}: {Optimizing} video encoding for video analytics.
\newblock In \emph{{MLSys}}, 2022.

\bibitem[He et~al.(2016)He, Zhang, Ren, and Sun]{He2016}
Kaiming He, Xiangyu Zhang, Shaoqing Ren, and Jian Sun.
\newblock Deep residual learning for image recognition.
\newblock In \emph{{CVPR}}, pages 770--778, 2016.

\bibitem[Hendrycks and Dietterich(2019)]{Hendrycks2018}
Dan Hendrycks and Thomas Dietterich.
\newblock Benchmarking neural network robustness to common corruptions and
  perturbations.
\newblock In \emph{{ICLR}}, 2019.

\bibitem[Howard et~al.(2019)Howard, Sandler, Chu, Chen, Chen, Tan, Wang, Zhu,
  Pang, Vasudevan, et~al.]{Howard2019}
Andrew Howard, Mark Sandler, Grace Chu, Liang-Chieh Chen, Bo Chen, Mingxing
  Tan, Weijun Wang, Yukun Zhu, Ruoming Pang, Vijay Vasudevan, et~al.
\newblock {Searching for MobileNetV3}.
\newblock In \emph{{ICCV}}, pages 1314--1324, 2019.

\bibitem[Hu et~al.(2023)Hu, Luo, Pasdar, Lee, Zhou, and Wu]{Hu2023}
Miao Hu, Zhenxiao Luo, Amirmohammad Pasdar, Young~Choon Lee, Yipeng Zhou, and
  Di Wu.
\newblock Edge-based video analytics: A survey.
\newblock \emph{{arXiv:2303.14329 [cs.DC]}}, 2023.

\bibitem[Hudson et~al.(2018)Hudson, L{\'e}ger, Niss, Sebesty{\'e}n, and
  Vaaben]{Hudson2018}
Graham Hudson, Alain L{\'e}ger, Birger Niss, Istv{\'a}n Sebesty{\'e}n, and
  J{\o}rgen Vaaben.
\newblock {JPEG-1 standard 25 years: past, present, and future reasons for a
  success}.
\newblock \emph{{J. Electron. Imaging}}, 27\penalty0 (4):\penalty0 040901,
  2018.

\bibitem[Isik et~al.(2023)Isik, Guleryuz, Tang, Taylor, and Chou]{Isik2023}
Berivan Isik, Onur~G. Guleryuz, Danhang Tang, Jonathan Taylor, and Philip~A.
  Chou.
\newblock Sandwiched video compression: {Efficiently} extending the reach of
  standard codecs with neural wrappers.
\newblock In \emph{ICIP}, pages 2055--2059, 2023.

\bibitem[Itsumi et~al.(2022)Itsumi, Beye, Charvi, and Nihei]{Itsumi2022}
Hayato Itsumi, Florian Beye, Vitthal Charvi, and Koichi Nihei.
\newblock Learning important regions via attention for video streaming on cloud
  robotics.
\newblock In \emph{{IROS}}, pages 6833--6839, 2022.

\bibitem[Kay et~al.(2017)Kay, Carreira, Simonyan, Zhang, Hillier,
  Vijayanarasimhan, Viola, Green, Back, Natsev, et~al.]{Kay2017}
Will Kay, Joao Carreira, Karen Simonyan, Brian Zhang, Chloe Hillier, Sudheendra
  Vijayanarasimhan, Fabio Viola, Tim Green, Trevor Back, Paul Natsev, et~al.
\newblock The {Kinetics} human action video dataset.
\newblock \emph{{arXiv:1705.06950 [cs.CV]}}, 2017.

\bibitem[Le et~al.(2021)Le, Zhang, Cricri, Ghaznavi-Youvalari, and
  Rahtu]{Le2021}
Nam Le, Honglei Zhang, Francesco Cricri, Ramin Ghaznavi-Youvalari, and Esa
  Rahtu.
\newblock Image coding for machines: an end-to-end learned approach.
\newblock In \emph{{ICASSP}}, pages 1590--1594, 2021.

\bibitem[Leclerc et~al.(2023)Leclerc, Ilyas, Engstrom, Park, Salman, and
  Madry]{Leclerc2023}
Guillaume Leclerc, Andrew Ilyas, Logan Engstrom, Sung~Min Park, Hadi Salman,
  and Aleksander Madry.
\newblock {FFCV}: {A}ccelerating training by removing data bottlenecks.
\newblock In \emph{{CVPR}}, pages 12011--12020, 2023.

\bibitem[Lederer(2019)]{Lederer2019}
Stefan Lederer.
\newblock 2019 video developer report – the future of video: {AV1} codec,
  {AI} \& machine learning, and low latency.
\newblock
  \url{https://bitmovin.com/bitmovin-2019-video-developer-report-av1-codec-ai-machine-learning-low-latency/},
  2019.

\bibitem[Li et~al.(2021)Li, Xie, Chen, Dollar, He, and Girshick]{Li2021}
Yanghao Li, Saining Xie, Xinlei Chen, Piotr Dollar, Kaiming He, and Ross
  Girshick.
\newblock Benchmarking detection transfer learning with vision transformers.
\newblock \emph{{arXiv:2111.11429 [cs.CV]}}, 2021.

\bibitem[Lin et~al.(2014)Lin, Maire, Belongie, Hays, Perona, Ramanan,
  Doll{\'a}r, and Zitnick]{Lin2014}
Tsung-Yi Lin, Michael Maire, Serge Belongie, James Hays, Pietro Perona, Deva
  Ramanan, Piotr Doll{\'a}r, and C~Lawrence Zitnick.
\newblock Microsoft {COCO}: {C}ommon objects in context.
\newblock In \emph{{ECCV}}, pages 740--755, 2014.

\bibitem[Liu et~al.(2021)Liu, Lin, Cao, Hu, Wei, Zhang, Lin, and Guo]{Liu2021b}
Ze Liu, Yutong Lin, Yue Cao, Han Hu, Yixuan Wei, Zheng Zhang, Stephen Lin, and
  Baining Guo.
\newblock {Swin Transformer}: {H}ierarchical vision transformer using shifted
  windows.
\newblock In \emph{{ICCV}}, pages 10012--10022, 2021.

\bibitem[Luo et~al.(2021)Luo, Talebi, Yang, Elad, and Milanfar]{Luo2021}
Xiyang Luo, Hossein Talebi, Feng Yang, Michael Elad, and Peyman Milanfar.
\newblock The rate-distortion-accuracy tradeoff: {JPEG} case study.
\newblock In \emph{{DCC}}, pages 354--354, 2021.

\bibitem[Mandhane et~al.(2022)Mandhane, Zhernov, Rauh, Gu, Wang, Xue, Shang,
  Pang, Claus, Chiang, et~al.]{Mandhane2022}
Amol Mandhane, Anton Zhernov, Maribeth Rauh, Chenjie Gu, Miaosen Wang, Flora
  Xue, Wendy Shang, Derek Pang, Rene Claus, Ching-Han Chiang, et~al.
\newblock {MuZero} with self-competition for rate control in {VP9} video
  compression.
\newblock \emph{{arXiv:2202.06626 [eess.IV]}}, 2022.

\bibitem[Menghani(2023)]{Menghani2023}
Gaurav Menghani.
\newblock Efficient deep learning: {A} survey on making deep learning models
  smaller, faster, and better.
\newblock \emph{{ACM Comput. Surv.}}, 55\penalty0 (12):\penalty0 1--37, 2023.

\bibitem[Mishra et~al.(2022)Mishra, Singh, and Singh]{Mishra2022}
Dipti Mishra, Satish~Kumar Singh, and Rajat~Kumar Singh.
\newblock Deep architectures for image compression: {A} critical review.
\newblock \emph{{Signal Process.}}, 191:\penalty0 108346, 2022.

\bibitem[{MMSegmentation Contributors}(2020)]{Mmseg2020}
{MMSegmentation Contributors}.
\newblock {MMSegmentation}: {OpenMMLab} semantic segmentation toolbox and
  benchmark.
\newblock \url{https://github.com/open-mmlab/mmsegmentation}, 2020.

\bibitem[Otani et~al.(2022)Otani, Hashiguchi, Omi, Fukushima, and
  Tamaki]{Otani2022}
Aoi Otani, Ryota Hashiguchi, Kazuki Omi, Norishige Fukushima, and Toru Tamaki.
\newblock Performance evaluation of action recognition models on low quality
  videos.
\newblock \emph{{IEEE Access}}, 10:\penalty0 94898--94907, 2022.

\bibitem[Park and Johnson(2023)]{Park2023}
Jeongsoo Park and Justin Johnson.
\newblock {RGB} no more: {M}inimally-decoded jpeg vision transformers.
\newblock In \emph{{CVPR}}, pages 22334--22346, 2023.

\bibitem[Reich et~al.(2023)Reich, Debnath, Patel, Prangemeier, and
  Chakradhar]{Reich2023}
Christoph Reich, Biplob Debnath, Deep Patel, Tim Prangemeier, and Srimat
  Chakradhar.
\newblock Deep video codec control.
\newblock \emph{{arXiv:2308.16215 [eess.IV]}}, 2023.

\bibitem[Reich et~al.(2024)Reich, Debnath, Patel, and Chakradhar]{Reich2024}
Christoph Reich, Biplob Debnath, Deep Patel, and Srimat Chakradhar.
\newblock Differentiable {JPEG}: The devil is in the details.
\newblock In \emph{{WACV}}, 2024.

\bibitem[Ren et~al.(2017)Ren, He, Girshick, and Sun]{Ren2017}
Shaoqing Ren, Kaiming He, Ross Girshick, and Jian Sun.
\newblock {Faster R-CNN}: {T}owards real-time object detection with region
  proposal networks.
\newblock \emph{{IEEE PAMI}}, 39\penalty0 (06):\penalty0 1137--1149, 2017.

\bibitem[Richardson(2004)]{Richardson2004}
Iain~E Richardson.
\newblock \emph{{H.264} and {MPEG-4} video compression: {Video} coding for
  next-generation multimedia}.
\newblock John Wiley \& Sons, 2004.

\bibitem[Russakovsky et~al.(2015)Russakovsky, Deng, Su, Krause, Satheesh, Ma,
  Huang, Karpathy, Khosla, Bernstein, Berg, and Fei-Fei]{Russakovsky2015}
Olga Russakovsky, Jia Deng, Hao Su, Jonathan Krause, Sanjeev Satheesh, Sean Ma,
  Zhiheng Huang, Andrej Karpathy, Aditya Khosla, Michael Bernstein,
  Alexander~C. Berg, and Li Fei-Fei.
\newblock {ImageNet large scale visual recognition challenge}.
\newblock \emph{{Int. J. Comp. Vis.}}, 115:\penalty0 211--252, 2015.

\bibitem[Shannon et~al.(1959)]{Shannon1959}
Claude~E Shannon et~al.
\newblock Coding theorems for a discrete source with a fidelity criterion.
\newblock \emph{{IRE Nat. Conv. Rec}}, 4\penalty0 (142-163):\penalty0 1, 1959.

\bibitem[Shin and Song(2017)]{Shin2017}
Richard Shin and Dawn Song.
\newblock {JPEG}-resistant adversarial images.
\newblock In \emph{{NIPS Workshop on Machine Learning and Computer Security}},
  2017.

\bibitem[Stone et~al.(2021)Stone, Maurer, Ayvaci, Angelova, and
  Jonschkowski]{Stone2021}
Austin Stone, Daniel Maurer, Alper Ayvaci, Anelia Angelova, and Rico
  Jonschkowski.
\newblock {SMURF}: {S}elf-teaching multi-frame unsupervised {RAFT} with
  full-image warping.
\newblock In \emph{{CVPR}}, pages 3887--3896, 2021.

\bibitem[Teed and Deng(2020)]{Teed2020}
Zachary Teed and Jia Deng.
\newblock {RAFT}: Recurrent all-pairs field transforms for optical flow.
\newblock In \emph{{ECCV}}, pages 402--419, 2020.

\bibitem[Tian et~al.(2021)Tian, Lu, Min, Che, Zhai, Guo, and Gao]{Tian2021}
Yuan Tian, Guo Lu, Xiongkuo Min, Zhaohui Che, Guangtao Zhai, Guodong Guo, and
  Zhiyong Gao.
\newblock {Self-conditioned probabilistic learning of video rescaling}.
\newblock In \emph{{ICCV}}, pages 4490--4499, 2021.

\bibitem[Tomar(2006)]{Tomar2006}
Suramya Tomar.
\newblock Converting video formats with {FFmpeg}.
\newblock \emph{{Linux J.}}, 146, 2006.

\bibitem[{TorchVision maintainers and contributors}(2016)]{Torchvision2016}
{TorchVision maintainers and contributors}.
\newblock {TorchVision}: {PyTorch's} computer vision library.
\newblock \url{https://github.com/pytorch/vision}, 2016.

\bibitem[Wallace(1992)]{Wallace1992}
Gregory~K Wallace.
\newblock {The JPEG still picture compression standard}.
\newblock \emph{{IEEE Trans. Consum. Electron.}}, 38\penalty0 (1):\penalty0
  xviii--xxxiv, 1992.

\bibitem[Wiegand et~al.(2003)Wiegand, Sullivan, Bjontegaard, and
  Luthra]{Wiegand2003}
Thomas Wiegand, Gary~J Sullivan, Gisle Bjontegaard, and Ajay Luthra.
\newblock Overview of the {H.264/AVC} video coding standard.
\newblock \emph{{IEEE Trans. Circuits Syst. Video Technol.}}, 13\penalty0
  (7):\penalty0 560--576, 2003.

\bibitem[Wightman(2019)]{Wightman2019}
Ross Wightman.
\newblock {PyTorch image models}.
\newblock \url{https://github.com/rwightman/pytorch-image-models}, 2019.

\bibitem[Xiao et~al.(2018)Xiao, Liu, Zhou, Jiang, and Sun]{Xiao2018}
Tete Xiao, Yingcheng Liu, Bolei Zhou, Yuning Jiang, and Jian Sun.
\newblock Unified perceptual parsing for scene understanding.
\newblock In \emph{{ECCV}}, pages 418--434, 2018.

\bibitem[Zhang et~al.(2023)Zhang, Zhu, Jiang, Kwong, and Kuo]{Zhang2023}
Yun Zhang, Linwei Zhu, Gangyi Jiang, Sam Kwong, and C.-C.~Jay Kuo.
\newblock {A survey on perceptually optimized video coding}.
\newblock \emph{{ACM Comput. Surv.}}, 55\penalty0 (12):\penalty0 1--37, 2023.

\end{thebibliography}
}

\end{document}